\documentclass{article}

\usepackage{graphicx}

\title{\textbf{Technical Report:} Sparse Hierachical Extrapolated Parametric Methods for Cortical Data Analysis
}

\begin{document}

\author{Nicolas Honnorat, Christos Davatzikos \\ Center for Biomedical Image Computing and Analytics \\ University of Pennsylvania, USA}

\maketitle
\begin{abstract}
	Many neuroimaging studies focus on the cortex, in order to benefit from better signal to noise ratios and reduced computational burden. Cortical data are usually projected onto a reference mesh, where subsequent analyses are carried out. Several multiscale approaches have been proposed for analyzing these surface data, such as spherical harmonics and graph wavelets. As far as we know, however, the hierarchical structure of the template icosahedral meshes used by most neuroimaging software has never been exploited for cortical data factorization. 
  In this paper, we demonstrate how the structure of the ubiquitous icosahedral meshes can be exploited by data factorization methods such as sparse dictionary learning, and we assess the optimization speed-up offered by extrapolation methods in this context. By testing different sparsity-inducing norms, extrapolation methods, and factorization schemes, we compare the performances of eleven methods for analyzing four datasets: two structural and two functional MRI datasets obtained by processing the data publicly available for the hundred unrelated subjects of the Human Connectome Project. Our results demonstrate that, depending on the level of details requested, a speedup of several orders of magnitudes can be obtained.

\end{abstract}

\section{Introduction}

Several modalities, such as EEG and MEG, are not able to image deep brain structures. 
MRI modalities benefit from better signal-to-noise ratios at the surface of the brain. 
For these practical reasons, many neuroimaging studies focus on the cortex. 
Cortical data are usually processed indepedently for the two hemispheres. 
The data of each hemisphere are projected onto a reference mesh, where subsequent analyses are carried out. 
Several multiscale approaches have been proposed for analyzing these surface data, such as spherical harmonics and graph wavelets \cite{wavelet_original,wavelet_golland,wavelet_chung}. As far as we know, however, these tools have not been exploited for accelerating data factorization schemes, such as 
nonnegative matrix factorization \cite{nnmf,pnmf2,enflure2} and sparse dictionary learning \cite{fista,palm}.

In this paper, we accelerate cortical data factorization by exploiting the hierarchical structure of icosahedral meshes 
and by investigating the effects of a novel extrapolation scheme adapted to nonnegative factorizations. 
Our results demonstrate that, depending on the level of details requested, a speedup of several orders of magnitudes can be obtained. 
The remainder of the paper is organized as follows. 
In section 2 we present the four factorization scheme considered for this work, we explain how mesh structure was exploited,  
how the factorizations were initialized and how they were accelerated by extrapolation. 
Section 3 presents our experimental results, obtained with two structural and functional datasets generated from the data available for the hundred unrelated HCP subjects \cite{hcp}. Discussions conclude the paper.

\section{Methods}

\subsection{Factorizations}

In this work, we compare four factorizing schemes: a variant of sparse dictionary learning \cite{palm}, 
two Non-negative Matrix factorizations (NNMF) \cite{nnmf,cichocki_book} and a projected Non-negative Matrix Factorization (PNMF) \cite{pnmf2}. 
Our goal is to decompose a data matrix $X$ of size $n_f \times n_s$, containing $n_f$ positive measurements acquired for $n_s$ subjects, 
as a product $BC$ between $n_d$ basis vectors, stored in a matrix $B$ of size $n_f \times n_d$ and loadings $C$ of size $n_d \times n_s$. 
We assume that the locations of the $n_f$ measures are in bijection with the faces of an icosahedral mesh ${\cal M}$ generated from the icosahedron ${\cal M}o$ as explained in the next section. 
In the context of neuroimaging the number of subjects $n_s$ is usually of the order of a few hundreds. 
The dimension $n_f$ is generally much larger, for instance 327680 faces for the largest HCP and freesurfer cortical meshes \cite{hcp,freesurfer}. 
This large dimension significantly slows down the computations and multiplies the local minima which could trap the alternating minimization 
scheme commonly used for solving factorization problems. 

In this work, we propose to reduce the spatial dimension by introducing a positive design matrix $D$ of size $n_f \times n_k$ and decomposing $X$ as follows:
\begin{eqnarray}
X \approx DBC
\end{eqnarray}
$D$ can be interpreted as an additional set of positive cortical basis. 
We start with a restricted number of coarse cortical maps, which we gradually refine 
to let the optimization focus on the cortical regions where decomposition errors are large.

For sparse dictionary learning, the L1 norm of the matrices $B$ and $C$ is penalized at the same time as the L2 decomposition error. 
Starting from a random initialization, $B$ and $C$ are iteratively updated by an alternating proximal gradient descent \cite{palm}.
Under the following notations
\begin{eqnarray}
L=X^{T} D ~~~~~,~~~~~ K=D^{T} D ~~~~~,~~~~~ M=L^{T} L 
\end{eqnarray}
the parametric dictionary learning problem solved and the alternating minimization scheme, known as PALM \cite{palm}, write: 
\begin{eqnarray*}
&(DL)& ~~~~\min{ \frac{1}{2}||X-DBC||_2^2 + \lambda\left(||B||_1+||C||_1\right) }\\
&~&\texttt{by repeating~~}
\left\{\begin{array}{l}
B \leftarrow S\left(B-\eta \left(KBCC^{T}-L^{T}C^{T}\right),\lambda \eta \right) \\
C \leftarrow S\left(C-\eta \left(B^{T}KBC-B^{T}L^{T}\right),\lambda \eta \right) 
\end{array}\right.
\end{eqnarray*}
where $\lambda$ is a constant sparsity parameter, we set $\eta=10^{-1}/||L||_2$, and the proximal operator $S$ applies a soft thresholding to matrix components:
\begin{eqnarray*}
S(z,\alpha)_{ij}=sign(z_{ij})\max{\left(0,|z_{ij}|-\alpha\right)}
\end{eqnarray*}

NNMF frameworks impose a positivity constraint on $B$ and $C$ and are usually optimized via multiplicative updates \cite{nnmf}. 
This constraint is often sufficient for generating sparse and non overlapping basis \cite{nnmf,pnmf2,cichocki_book}.
We compared two ``parametric'' NNMF frameworks: a framework where the square of the Frobenius norm of $B$ and $C$ is penalized to 
alleviate the ambiguity problem \cite{cichocki_book} and a framework generating sparse $B$ and $C$ \cite{cichocki_book}:
\begin{eqnarray*}
&(PNNMF)& ~~~~\min{ ||X-DBC||_2^2 + \lambda\left(||B||^2_2+||C||^2_2\right) }~~~~ B \geq 0 ~~ C \geq 0\\
&~&\texttt{by repeating~~}
\left\{\begin{array}{l}
B \leftarrow B \odot \left[L^{T}C^{T}-\lambda B\right]_{+} \oslash \left[K B C C^{T}\right] \\
C \leftarrow C \odot \left[B^{T}L^{T}-\lambda C\right]_{+} \oslash \left[B^{T} K B C\right] \\
\end{array}\right.~\\
&(SPNNMF)& ~~~~\min{ \frac{1}{2}||X-DBC||_2^2 + \lambda\left(||B||_1+||C||_1\right) }~~~~ B \geq 0 ~~ C \geq 0\\
&~&\texttt{by repeating~~}
\left\{\begin{array}{l}
B \leftarrow B \odot \left[L^{T}C^{T}-\lambda \mathbf{1}(n_k,n_d)\right]_{+} \oslash \left[K B C C^{T}\right] \\
C \leftarrow C \odot \left[B^{T}L^{T}-\lambda \mathbf{1}(n_d,n_s)\right]_{+} \oslash \left[B^{T} K B C\right]
\end{array}\right.
\end{eqnarray*}
where $\odot$ denotes the entrywise product, $\oslash$ the entrywise division, $\left[x\right]_{+}=max(x,0)$ and $\mathbf{1}(n,m)$ is the matrix of ones of size as $n \times m$.

In addition, we considered a projective NMF (PNMF) scheme \cite{pnmf2}. 
PNMF generates the loadings by projecting the data on the basis. As a result, only $B$ needs to be determined and no penalization is required for handling ambiguity issues. We obtained the following parametric scheme after reducing the amplitude of the updates by two for stabilizing the optimization \cite{cichocki_book}:
\begin{eqnarray*}
(PPNMF) ~~~~\min{ ||X-DBB^{T}D^{T}X||_2^2 } ~~~~~~~~ B \geq 0 ~~~~~~~~~~~~~~~~~~~~~~~~\\
\texttt{by repeating}~~~
B \leftarrow B \odot \left(\frac{\mathbf{1}(n_k,n_d)}{2}+\left[MB\right] \oslash \left[ \left(KBB^{T}M+MBB^{T}K\right)B \right]\right) \\
\end{eqnarray*}

\begin{figure}[t]
\includegraphics[height=0.2\linewidth]{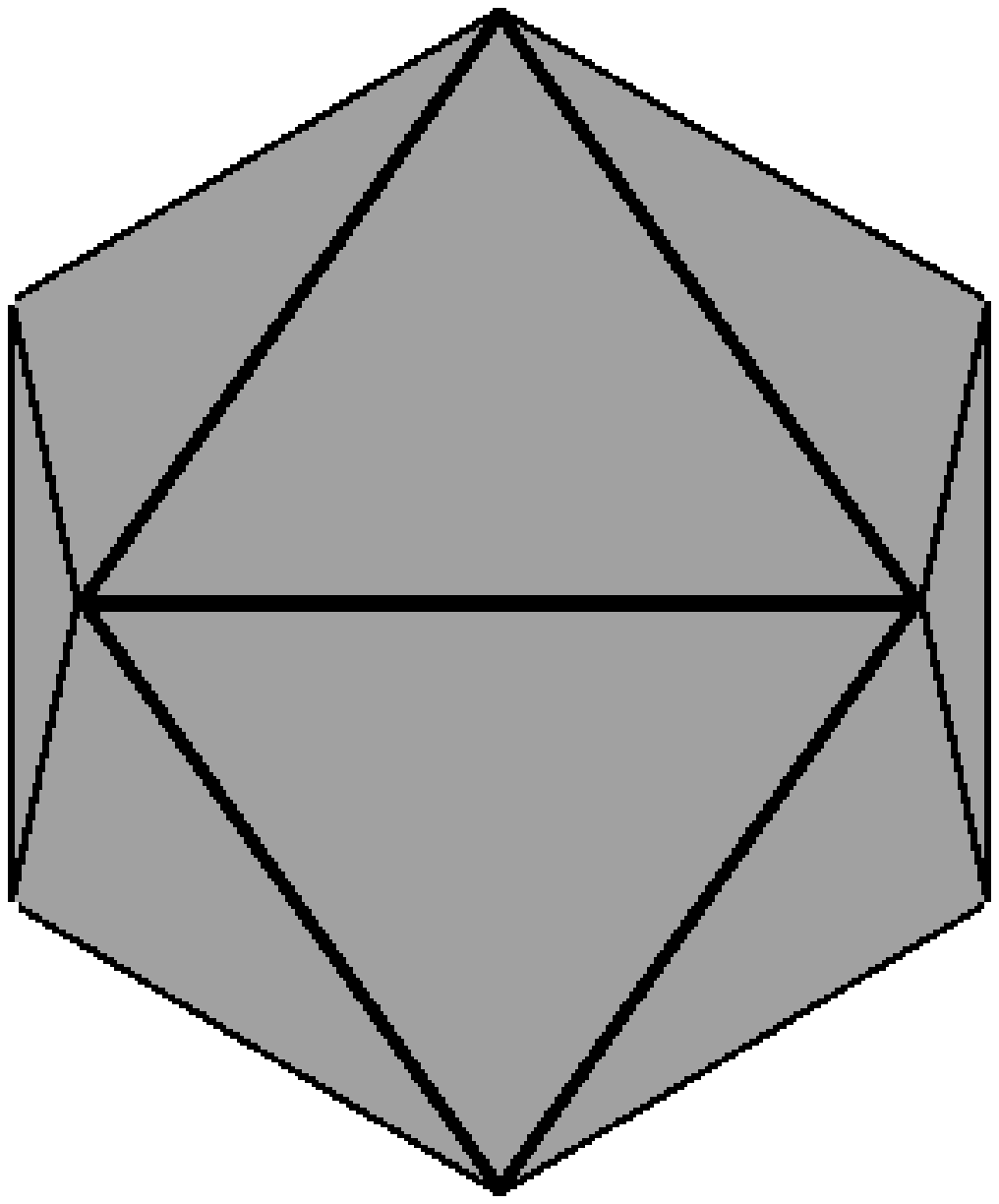}
\includegraphics[height=0.125\linewidth]{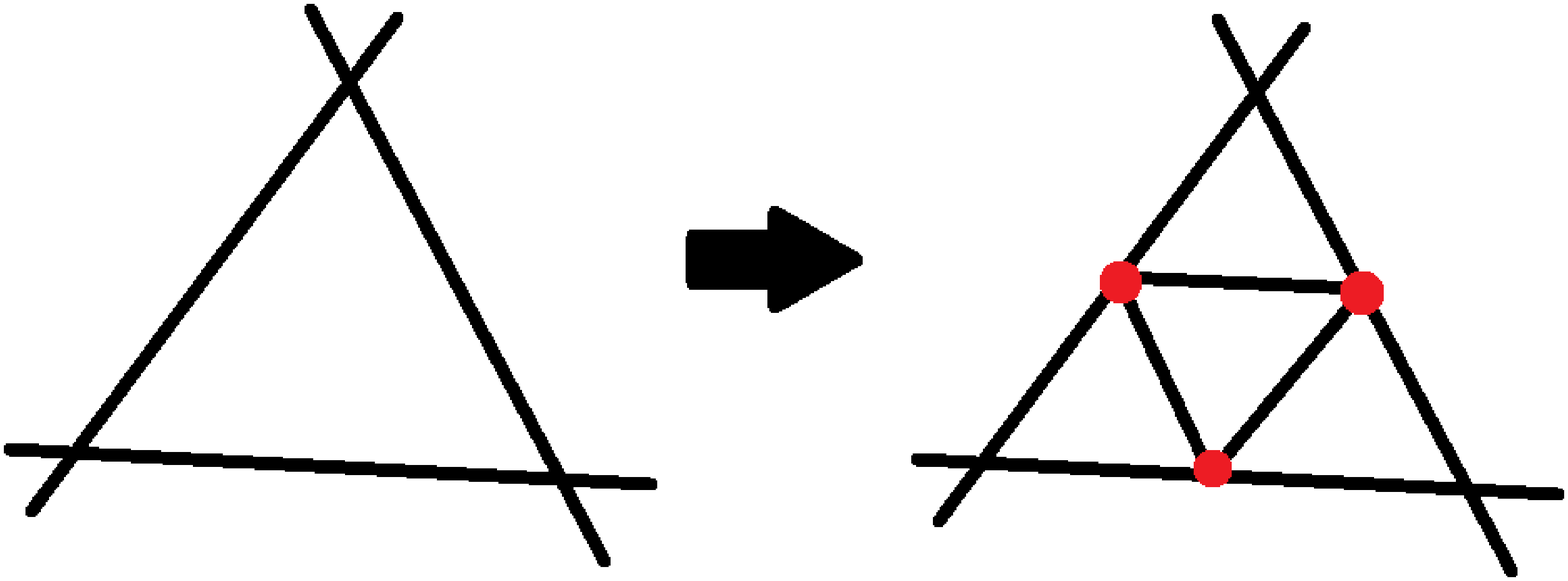}
\includegraphics[height=0.2\linewidth]{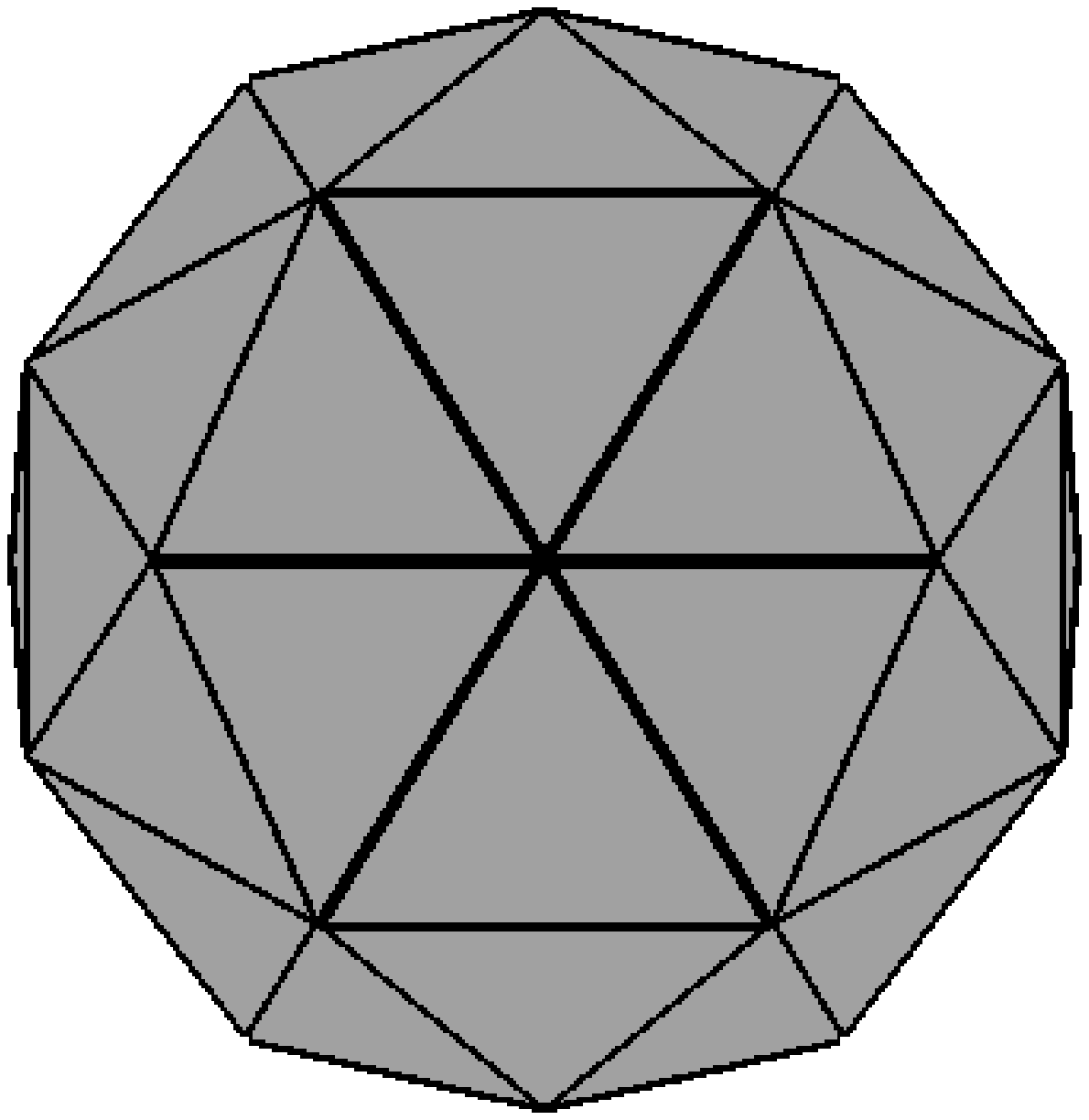}
~~~~\raisebox{4em}{
$\begin{array}{|c|c|}
\hline
~ & \texttt{level~} n\\
\hline
\texttt{faces} & 20 \times 4^n \\
\hline
\texttt{edges} & 30 \times 4^n \\
\hline
\texttt{nodes} & 10 \times 4^n +2 \\
\hline
\end{array}$
}
\caption{\label{methode:icosahedrons} Icosahedral meshes are generated by iteratively dividing the faces of an icosahedron and projecting the 
new nodes on the unit sphere. The numbers of faces, edges, and nodes on the right are obtained when dividing by four as illustrated here.}
\end{figure}

\subsection{Hierarchical Optimization on Icosahedral Meshes}

Icosahedral meshes are generated by iteratively subdividing the faces of an icosahedron in four smaller triangles, as illustrated in figure \ref{methode:icosahedrons}. 
This procedure generates almost perfectly regular meshes. 
Spherical harmonics and graph wavelets allow to exploit the hierarchical structure of these meshes \cite{wavelet_original,wavelet_golland,wavelet_chung}. 
In order to preserve positivity when factorizing cortical data, we adopted an approach more primitive but closely related to spherical wavelets \cite{wavelet_original}. 
More precisely, we initialize $D$ by concatenating twenty rotated versions of the same positive cortical map, centered on the faces of the original icosahedron. 
A thousand time, we initialize $B$ and $C$ randomly and run a thousand iterations of our factorization method. The best pair $(B,C)$ is then gradually refined by a procedure preserving the association between the column of D and the faces of a icosahedral mesh gradually refined. 
Each refinement consists of two steps. 
An average local error is first computed for each column of D by projecting the reconstruction error $X-DBC$ using $D$:
\begin{eqnarray}
e=\left[\left[L^{T}-KBC\right] \odot \left[L^{T}-KBC\right] \right]\mathbf{1}(n_s,1)
\end{eqnarray}
Then, the faces associated with the worst errors are subdivided, D is updated by rotating the removed columns and scaling their support by half accordingly, the matrices $K,L,M$ required for the factorization are updated, and $B$ is refined by dividing by four and replicating four time the rows corresponding to the columns removed from D. 
Local errors $e$ and $B$ updates are not exact because $D$ is not an orthogonal basis. 
However, these updates are very efficient and constitutes good approaximations when the overlap between the cortical maps in $D$ is limited.

\begin{figure}[t]
\scriptsize
\begin{tabular}{|l|c|c|}
\hline
(PNNMF) standard & extrapolation & bounded log extrapolation (LE)\\
\hline
\parbox{4.0cm}{
\begin{eqnarray*}
\Delta^{+}&=& \left[L^{T}C^{T}-\lambda B\right]_{+}  \\
\Delta^{-}&=& K B C C^{T}\\
B^{t+1} &\leftarrow& B^{t} \odot \Delta^{+} \oslash \Delta^{-}
\end{eqnarray*}
}  &
\parbox{5.0cm}{
\begin{eqnarray*}
y^{t} &\leftarrow& B^{t} \odot \Delta^{+} \oslash \Delta^{-} \\
B^{t+1} &\leftarrow& \left[ y^{t}+\frac{\tau-1}{\tau+1}(y^{t}-y^{t-1}) \right]_{+} \\
\tau &\leftarrow& \frac{1+\sqrt{1+4\tau^2}}{2}
\end{eqnarray*}
}  &  
\parbox{5.5cm}{
\begin{eqnarray*}
y^{t} &\leftarrow& B^{t} \odot \Delta^{+} \oslash \Delta^{-} \\
B^{t+1} &\leftarrow& y^{t} \odot \left[\left[y^{t} \oslash y^{t-1}\right]^{\frac{\tau-1}{\tau+1}} \right]_{0.1}^{10} \\
\tau &\leftarrow& \frac{1+\sqrt{1+4\tau^2}}{2}
\end{eqnarray*}
}\\
\hline
\end{tabular}
\normalsize
\caption{Proximal gradient updates for $B$ for the PNNMF schemes, with notation $\left[x\right]_{0.1}^{10}=min(10,max(0.1,x))$. $\tau$ starts at $\frac{1+\sqrt{5}}{2}$ \cite{fista}. LE power is taken componentwise.}
\label{extrapolation}
\end{figure}

\subsection{Initialization and Design Matrices}

For dictionary learning, $B$ and $C$ were initialized by sampling from ${\cal N}(0,1)$. 
For PPNMF schemes, $B$ was then scaled by the inverse of the Frobenius norm of $L$ and replaced by its absolute value. 
For NMF schemes, $C$ was initialized uniformly with $\frac{1}{n_d}\mathbf{1}(n_d,n_s)$ and $B$ was generated by averaging
random selections of five colums of $L$\cite{nmf_init}. 
The following positive functions
\begin{eqnarray*}
f(x)= \left\{
\begin{array}{ll}
exp\left( -\frac{ {cos}^{-1}(\langle x_o,x \rangle)}{\pi \sigma} \right) & ~~~~~\frac{{cos}^{-1}(\langle x_o,x \rangle)}{\pi \sigma} \leq \tau \\
0 & ~~~~~\texttt{otherwise}
\end{array}\right.
\end{eqnarray*}
where $\langle .,. \rangle$ is the standard inner product, were evaluated for $(\sigma=0.015,\tau=3)$, for each of the $n_f$ face centers $x$ of ${\cal M}$ and for $x_o$ center of the twenty original ${\cal M}o$ icosahedron faces. We built an initialization $D_O$ for $D$ by concatenating these twenty cortical maps, which means that optimization started at $n_k=20$.

\subsection{Extrapolation}

Extrapolation was introduced for accelerating the convergence of iterative soft-thresholding algorithms \cite{fista}. 
Extrapolation exploits previous gradient steps for extending the current one, as shown in figure \ref{extrapolation}. 
For NMF schemes, an additional projection is necessary to preserve positivity. 
However, this projection generates zeros which can trap the optimization. 
For this reason, we investigated the log extrapolation illustrated in figure \ref{extrapolation}. 
This novel extrapolation, which was bounded in amplitude and run after ten standard updates to prevent instability, 
corresponds to an extrapolation of the logarithm of the matrix components.

\section{Results}
\begin{figure}[t]
(1)\includegraphics[width=0.45\linewidth]{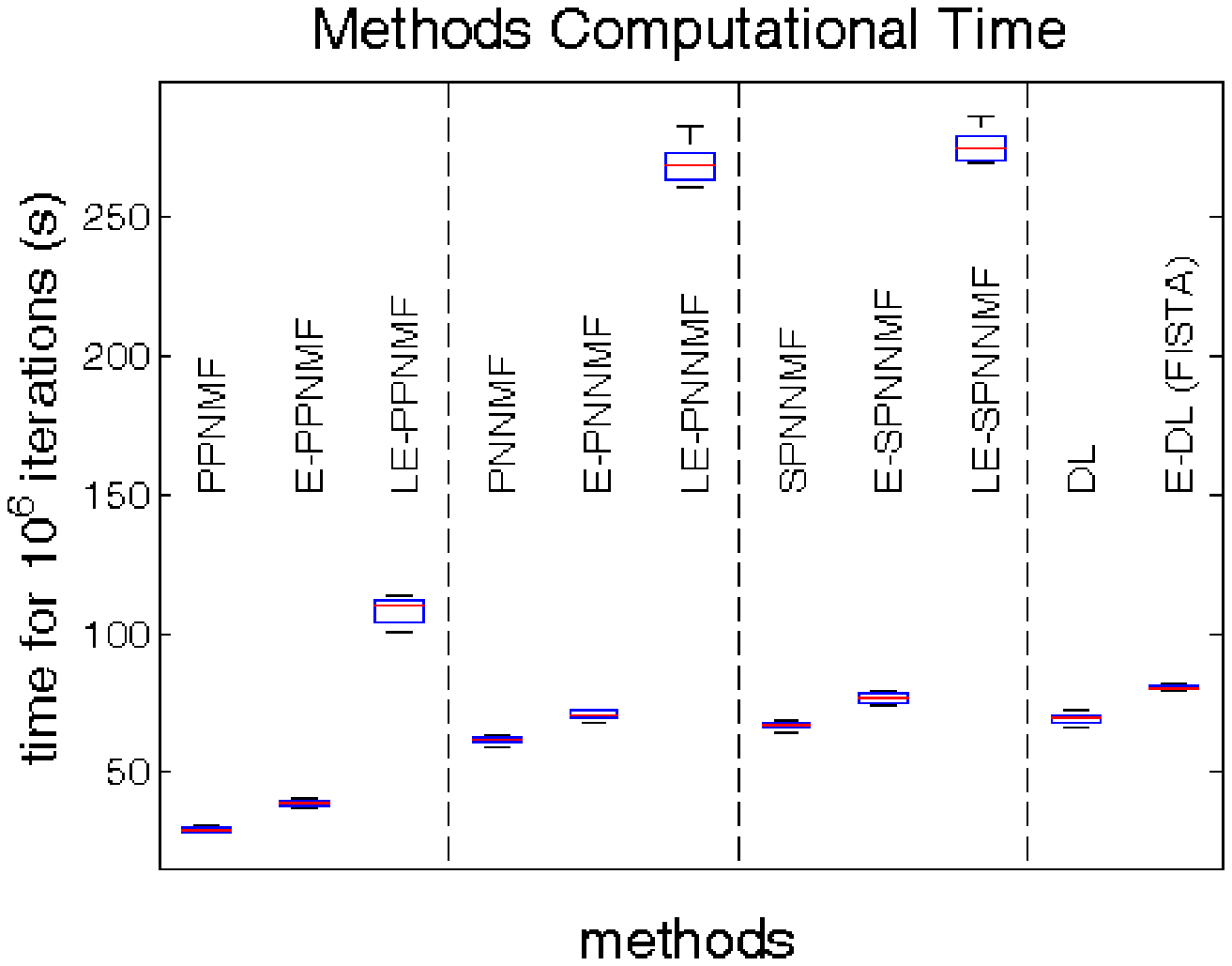}
(2)\includegraphics[width=0.45\linewidth]{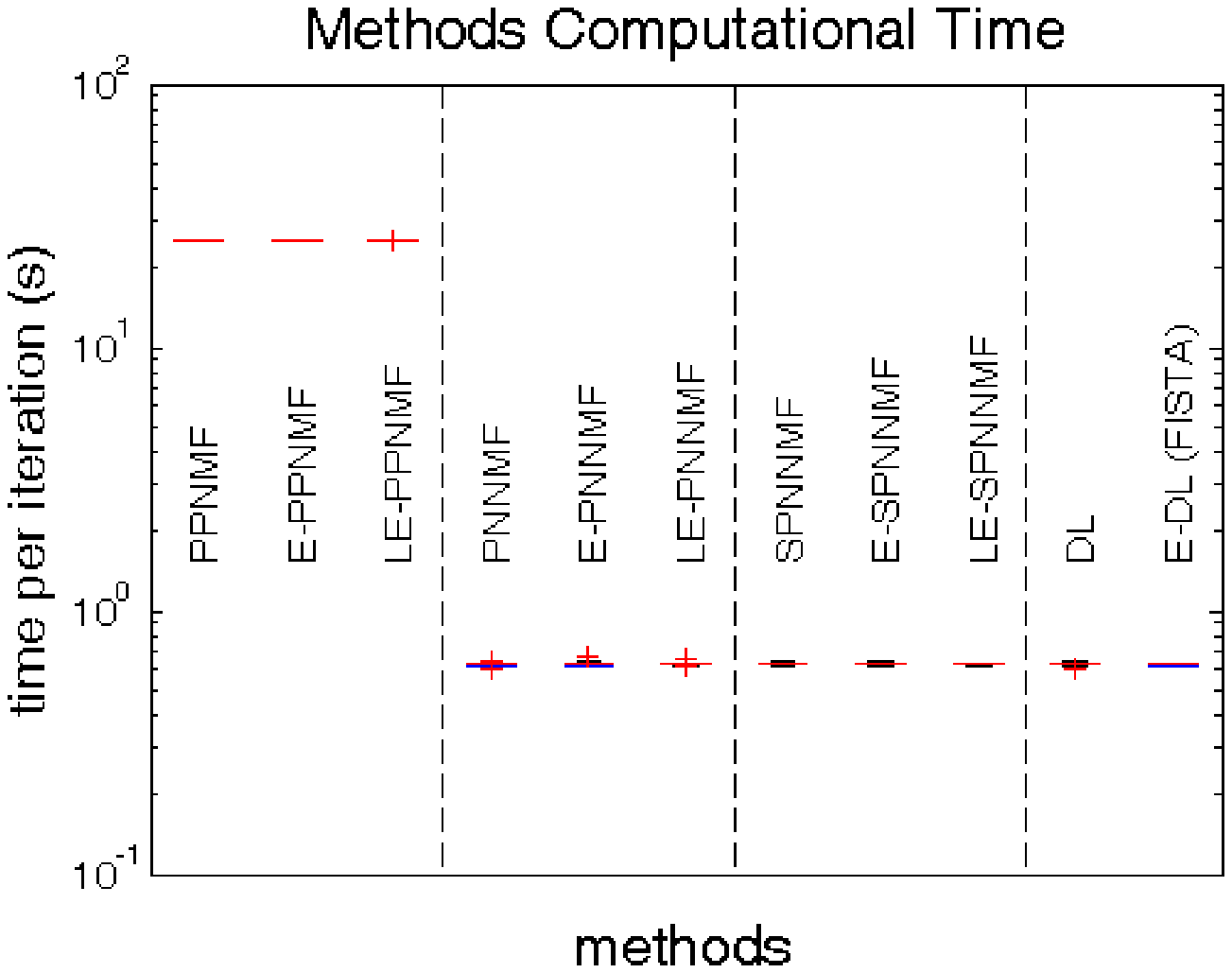}
\caption{(1) Time required for a million iterations for $D=D_O,~n_d=10$ (10 repetitions). (2) Time for a single iteration at original mesh resolution (10 242 nodes,10 repetitions).}
\label{res:time}
\end{figure}

\subsection{Data and Parameters}

We validated our methods using the data of the hundred unrelated HCP subjects \cite{hcp}. The outliers of each subject myelin and cortical thickness maps were removed by limiting the absolute difference with the median to 4.4478 median absolute deviations. 
We generated a regional homogeneity map (reHo) \cite{reho} for each subject by first bandpass-filtering between 0.05 and 0.1 Hz the rs-fMRI processed with the ICA+FIX pipeline with MSMAll registration \cite{hcp_preprocessing}. The timeseries obtained were then normalized to zero mean and unit L2 norm, concatenated, and reHo was measured for neighborhoods of three edges of radius \cite{reho}. The amplitude of low frequency fluctuations (ALFF) \cite{alff} was on the contrary measured for each subject scan separately, for the frequency band 0.05-0.1 Hz. All these positive maps were projected onto the fsaverage5 mesh using the transformation provided on Caret website \cite{caret}. For all the experiments sparsity level $\lambda$ was set to $5$ for dictionary learning, $1/2$ for sparse NMF schemes. 
$\lambda$ was set to $1/||L||_2$ for alleviating ambuiguity issues with non-sparse NMF schemes.

\subsection{Computational Time and Extrapolation}

We measured the maximal speed up by comparing the computational time required for running the first million of optimization steps with the small design matrix $D_O$ with the computational time required for running the same algorithm at the original fsaverage5 resolution. 
The results presented in figure \ref{res:time} correspond to speedups between $2.4 \times 10^3$ and $6.25 \times 10^5$ for the most time consuming projected NMF schemes. 
These speedups of three to six orders of magnitude grant us the possibility to generate a good initial pair $(B,C)$ by running a large number of random initializations.

\begin{figure}[t]
\includegraphics[width=0.475\linewidth]{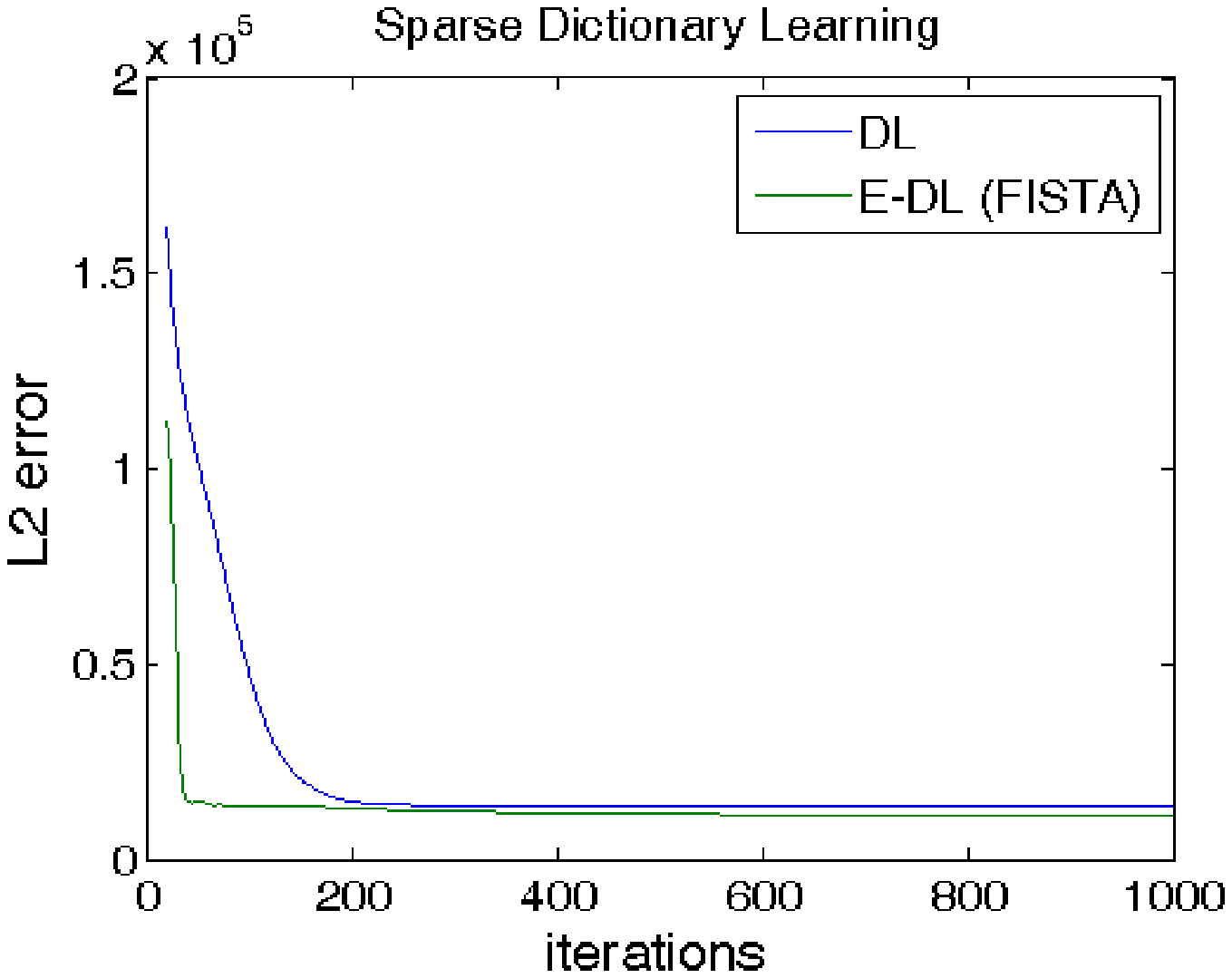}
\includegraphics[width=0.475\linewidth]{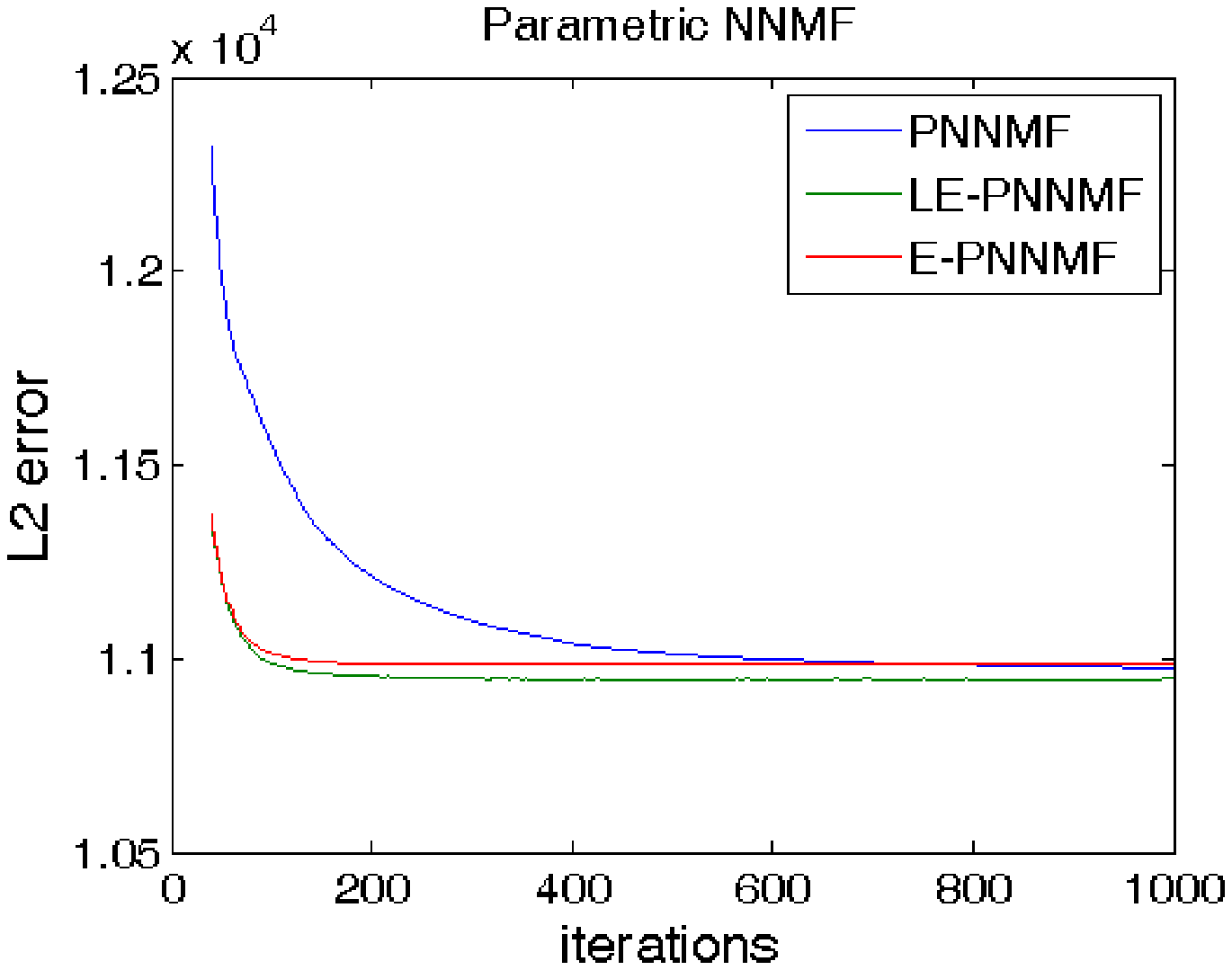} \\
\includegraphics[width=0.475\linewidth]{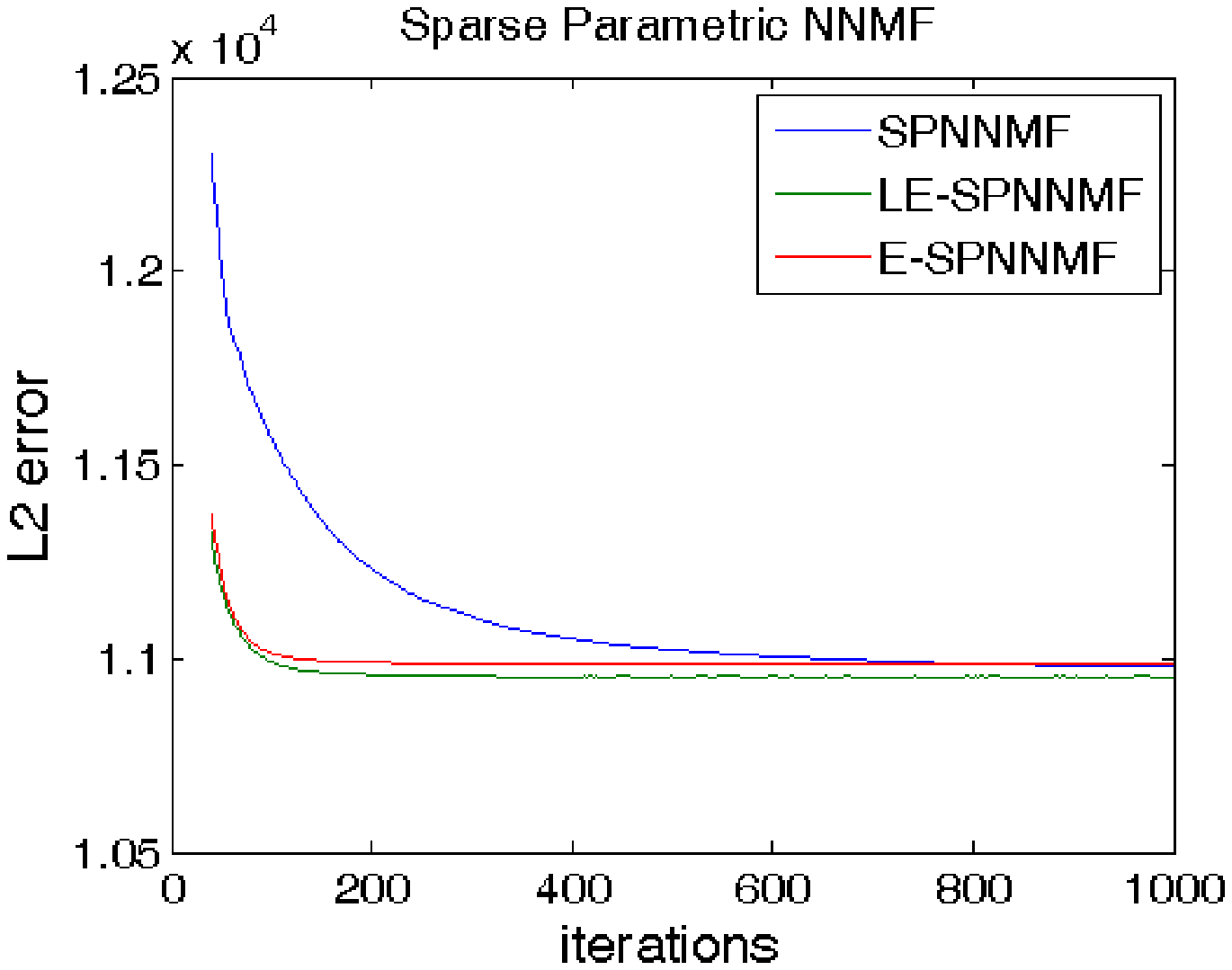}
\includegraphics[width=0.475\linewidth]{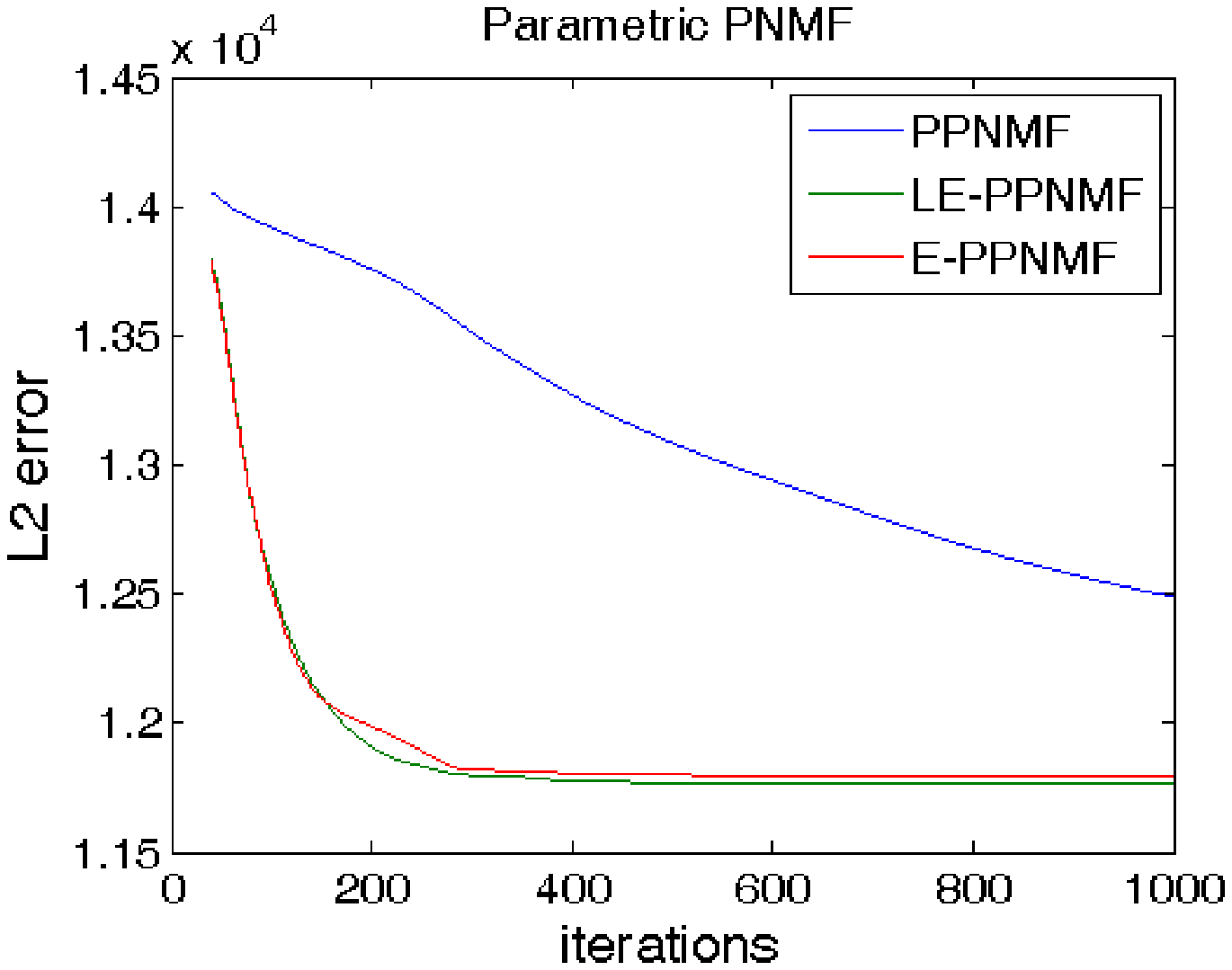}
\caption{Factorization schemes convergence, with standard extrapolation (E-) or log-extrapolation (LE-) for left hemisphere myelin data. We report the median of ten runs.}
\label{res:cvg}
\end{figure}

We compared the extrapolation strategies by measuring the reconstruction error $||X-DBC||_2^2$ when running our eleven algorithms ten times for factorizing the left hemisphere myelin data. $D$ was set to $D_O$ and the algorithms were run for a thousand iterations. The median errors reported in figure \ref{res:cvg} demonstrate that both extrapolation approaches significantly speed up the convergence. However, the log-extrapolation is less likely to get trapped in local minima, and always reaches a slightly lower energy than state of the art extrapolation. 

\subsection{Methods Comparison}

\begin{figure}[t]
\includegraphics[width=0.35\linewidth]{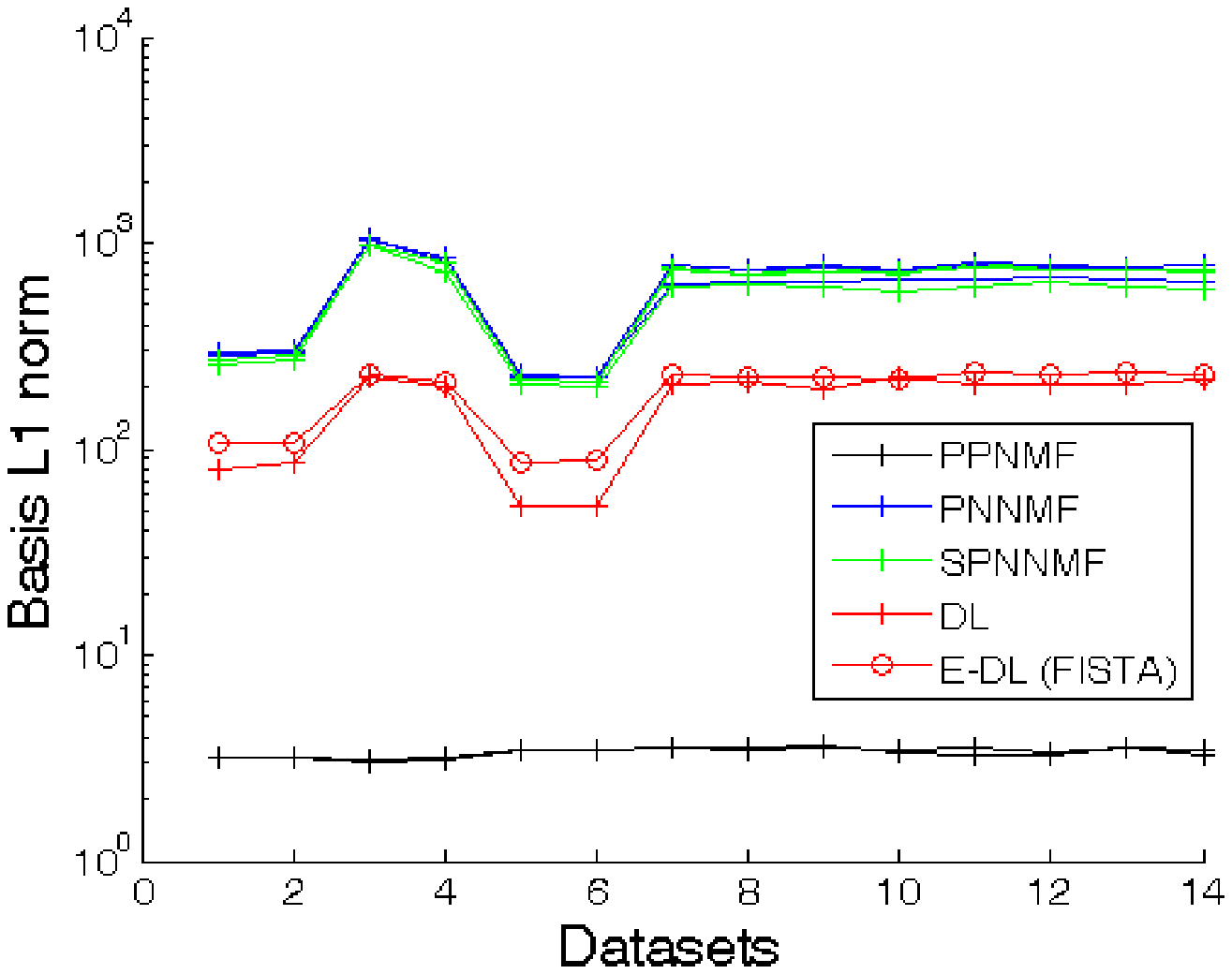}
\includegraphics[width=0.35\linewidth]{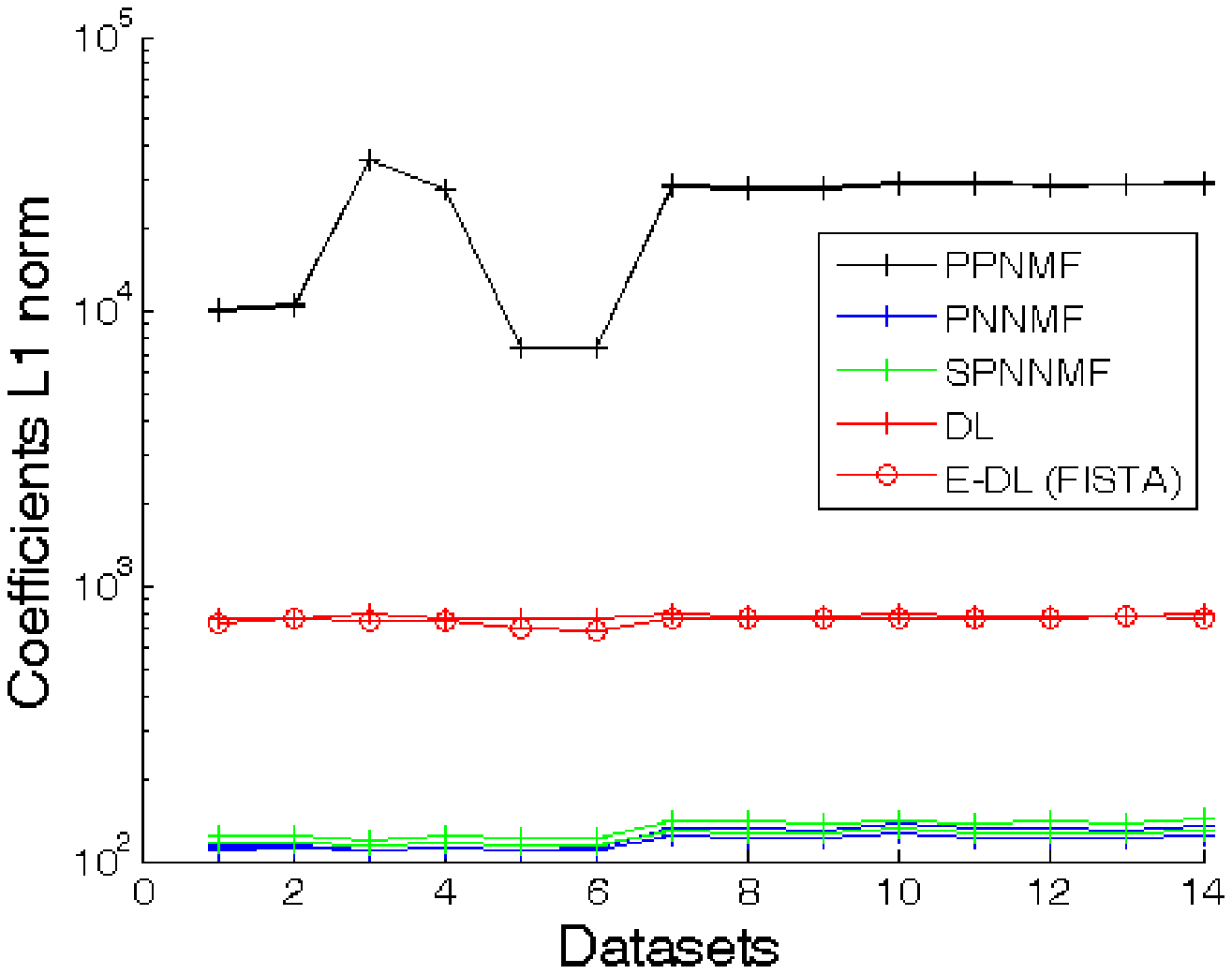} 
\raisebox{5em}{
\scriptsize{\begin{tabular}{|l|l|}
\hline
 & DATASETS \\
\hline
1,2 & myelin (Left,Right) \\
\hline
3,4 & cortical thickness (L.,R.) \\
\hline
5,6 & reHo (L.,R.) \\
\hline
7,8 & ALFF, scan 1 (L.,R.) \\
\hline
9,10 & ALFF, scan 2 (L.,R.) \\
\hline
11,12 & ALFF, scan 3 (L.,R.) \\
\hline
13,14 & ALFF, scan 4 (L.,R.) \\
\hline
\end{tabular}}
}
\normalsize
\caption{L1 norms of the basis B and loadings C generated by all the algorithms tested.}
\label{res:final}
\end{figure}

\begin{figure}[t]
\includegraphics[width=0.22\linewidth]{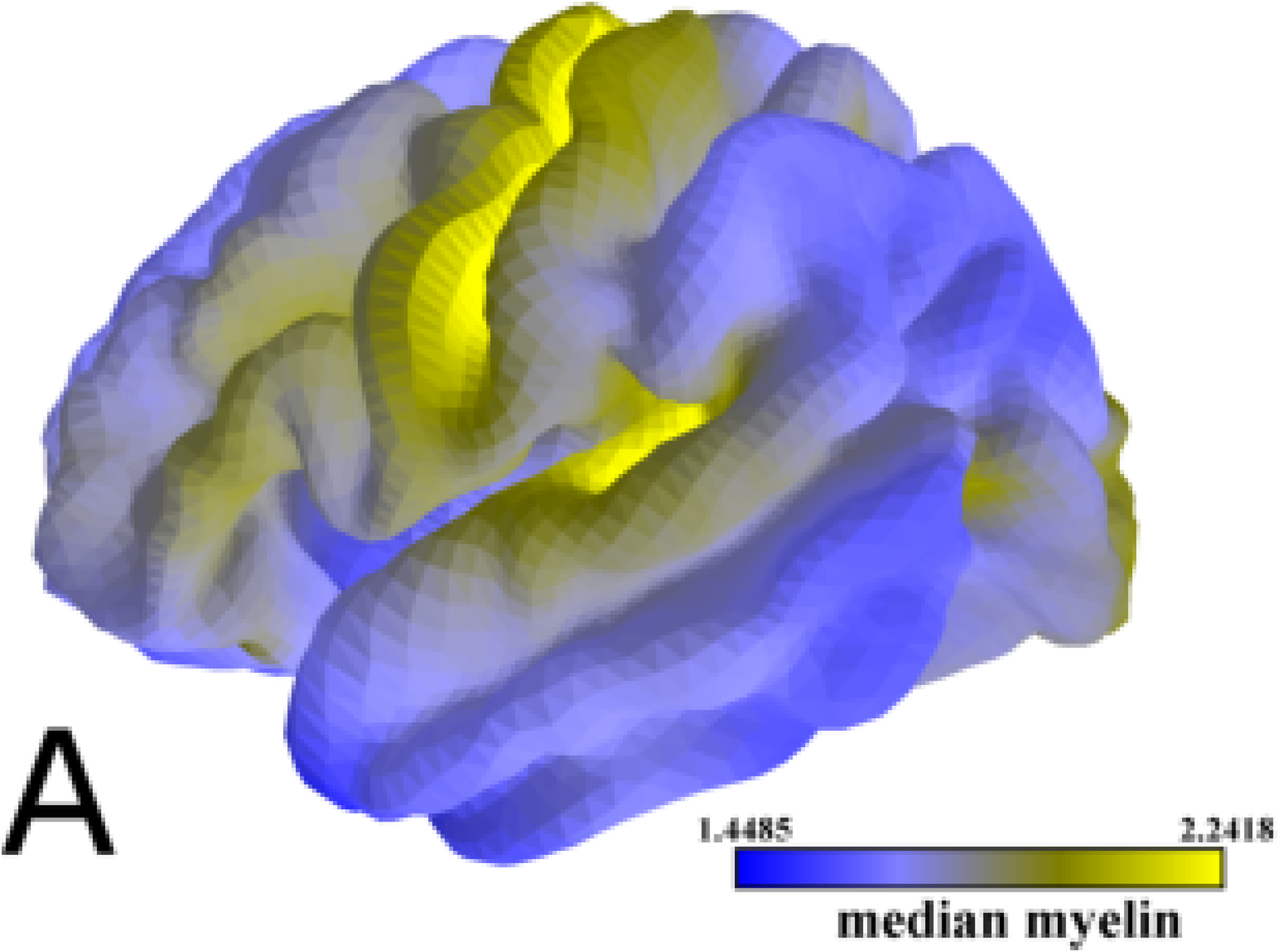}
\includegraphics[width=0.22\linewidth]{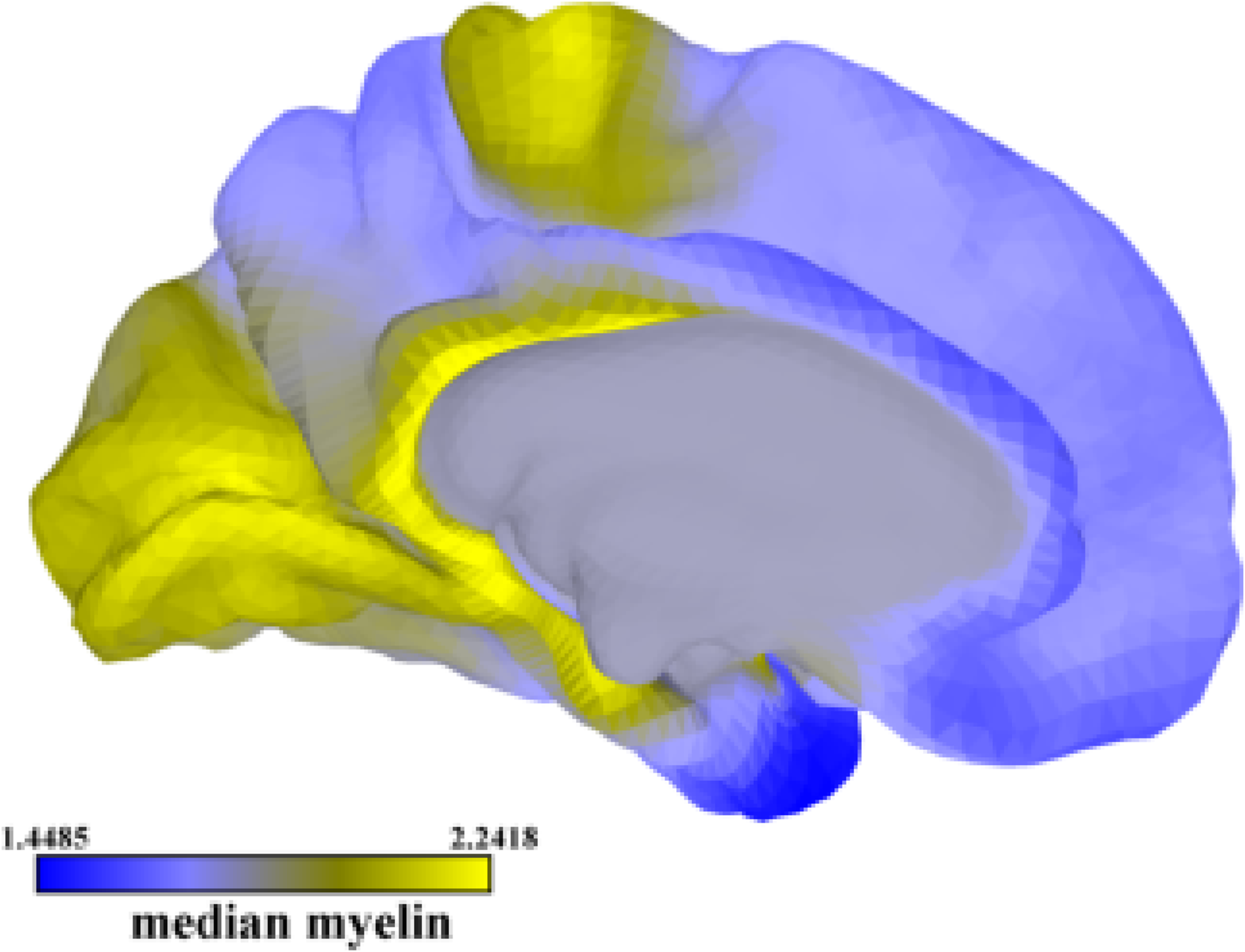}
\includegraphics[width=0.22\linewidth]{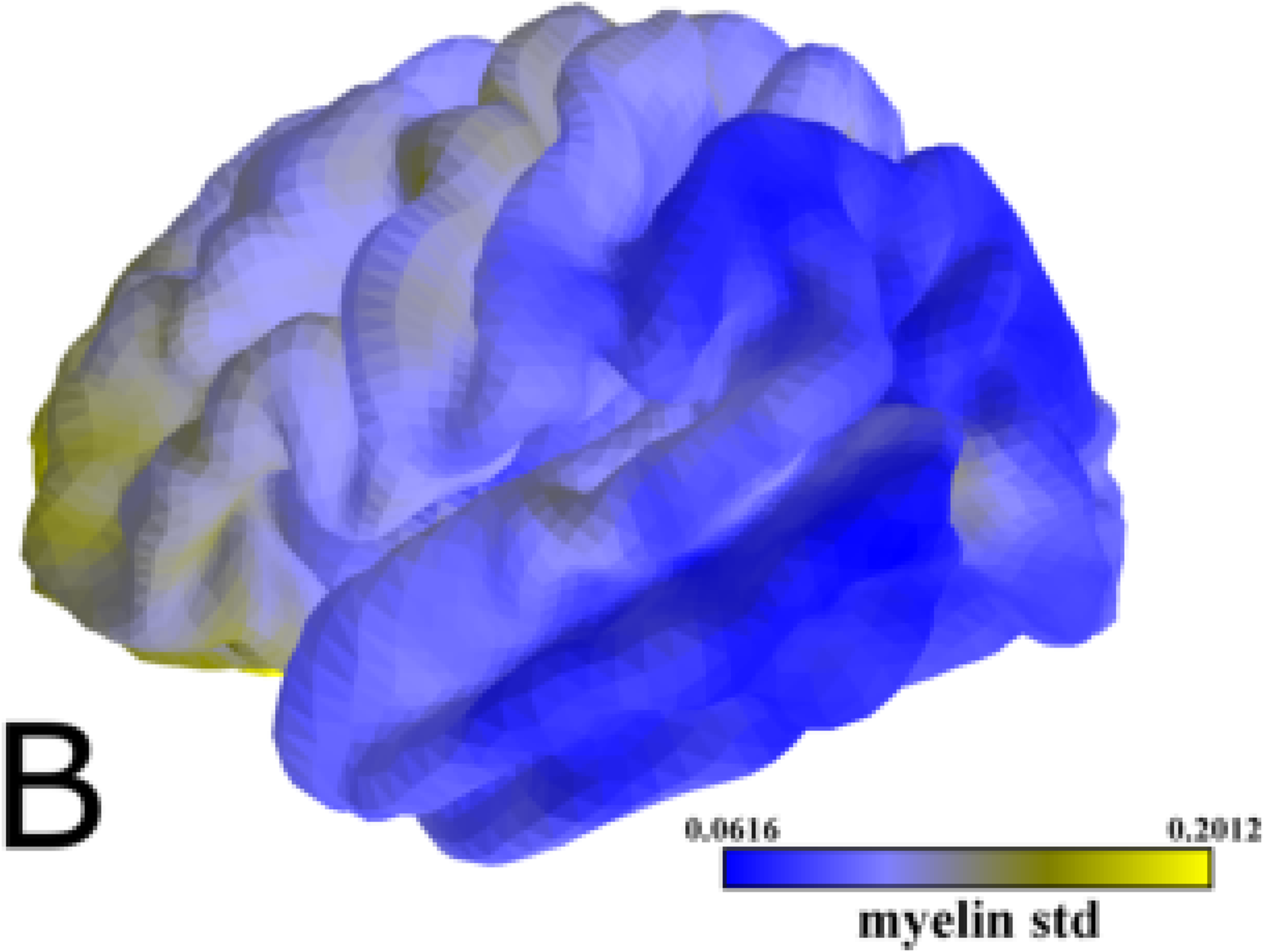}
\includegraphics[width=0.22\linewidth]{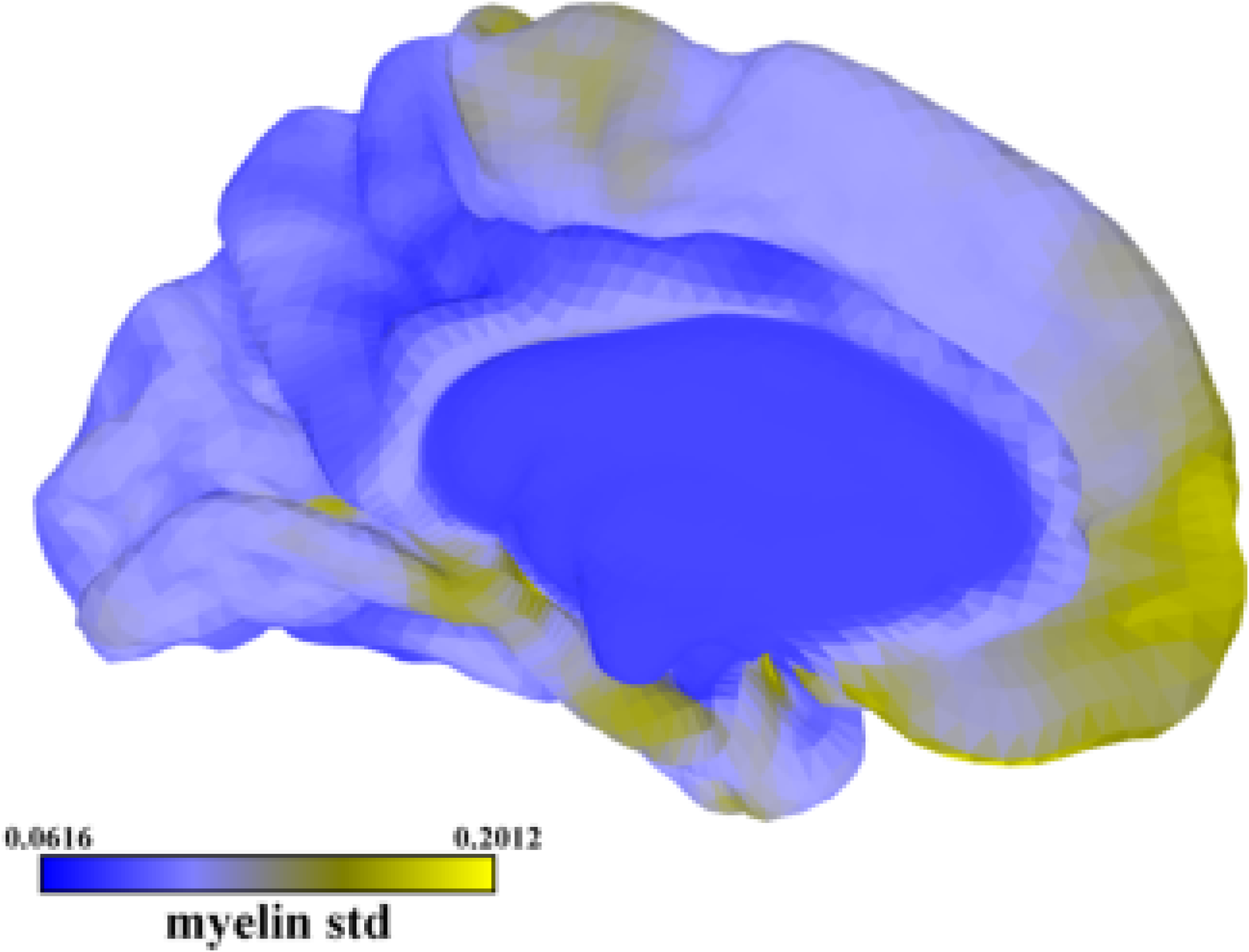} \\
\includegraphics[width=0.22\linewidth]{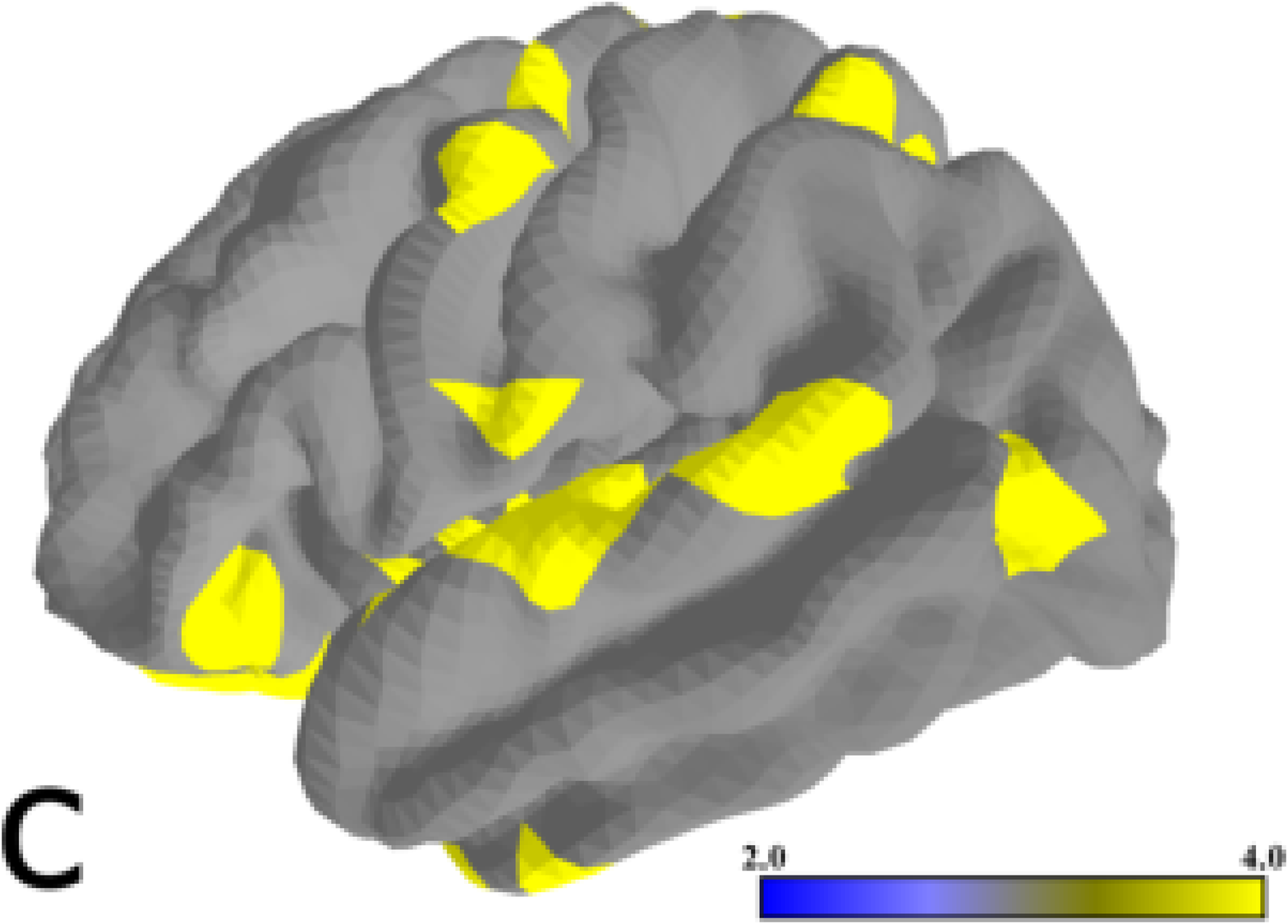}
\includegraphics[width=0.22\linewidth]{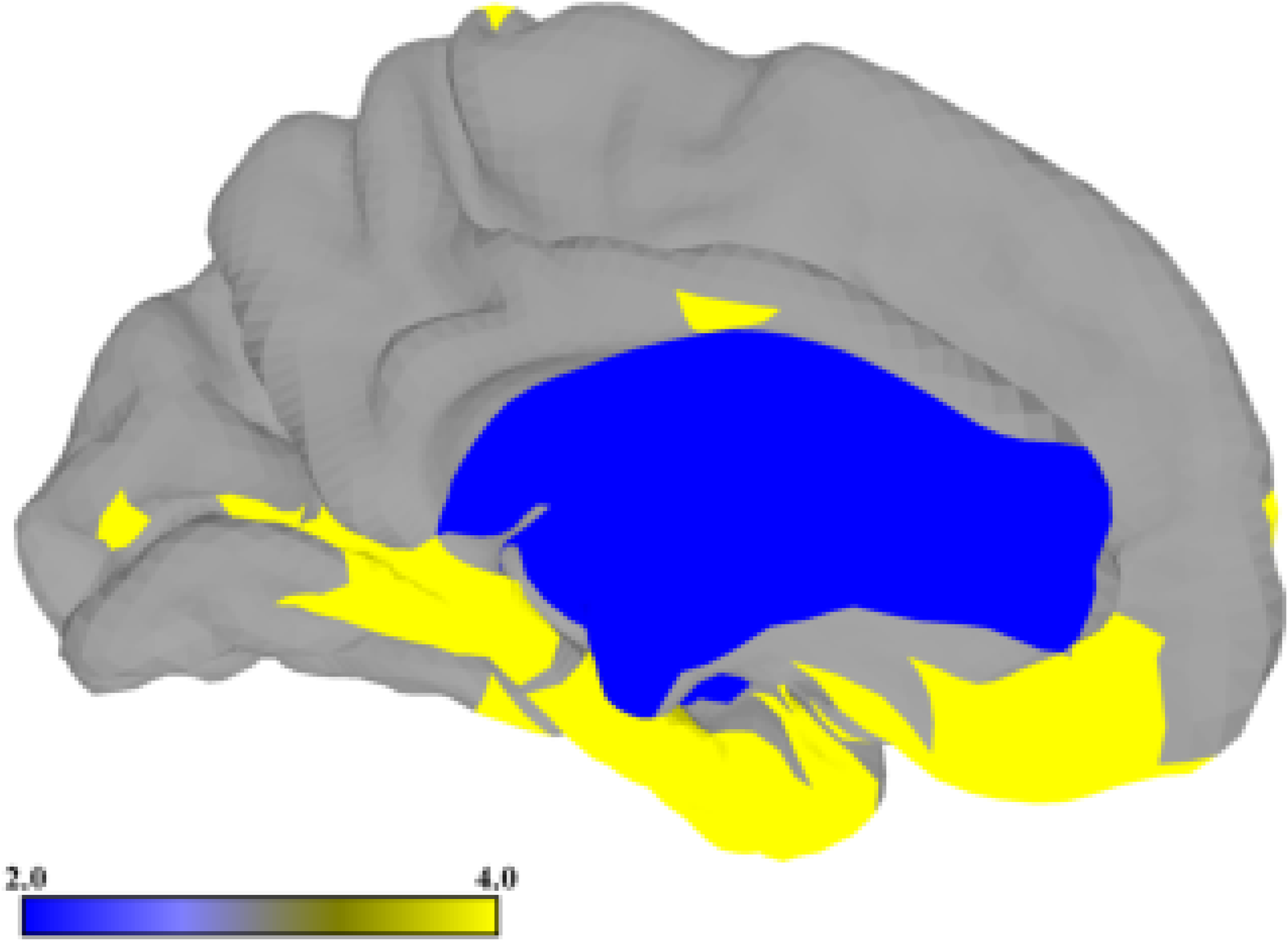}
\includegraphics[width=0.22\linewidth]{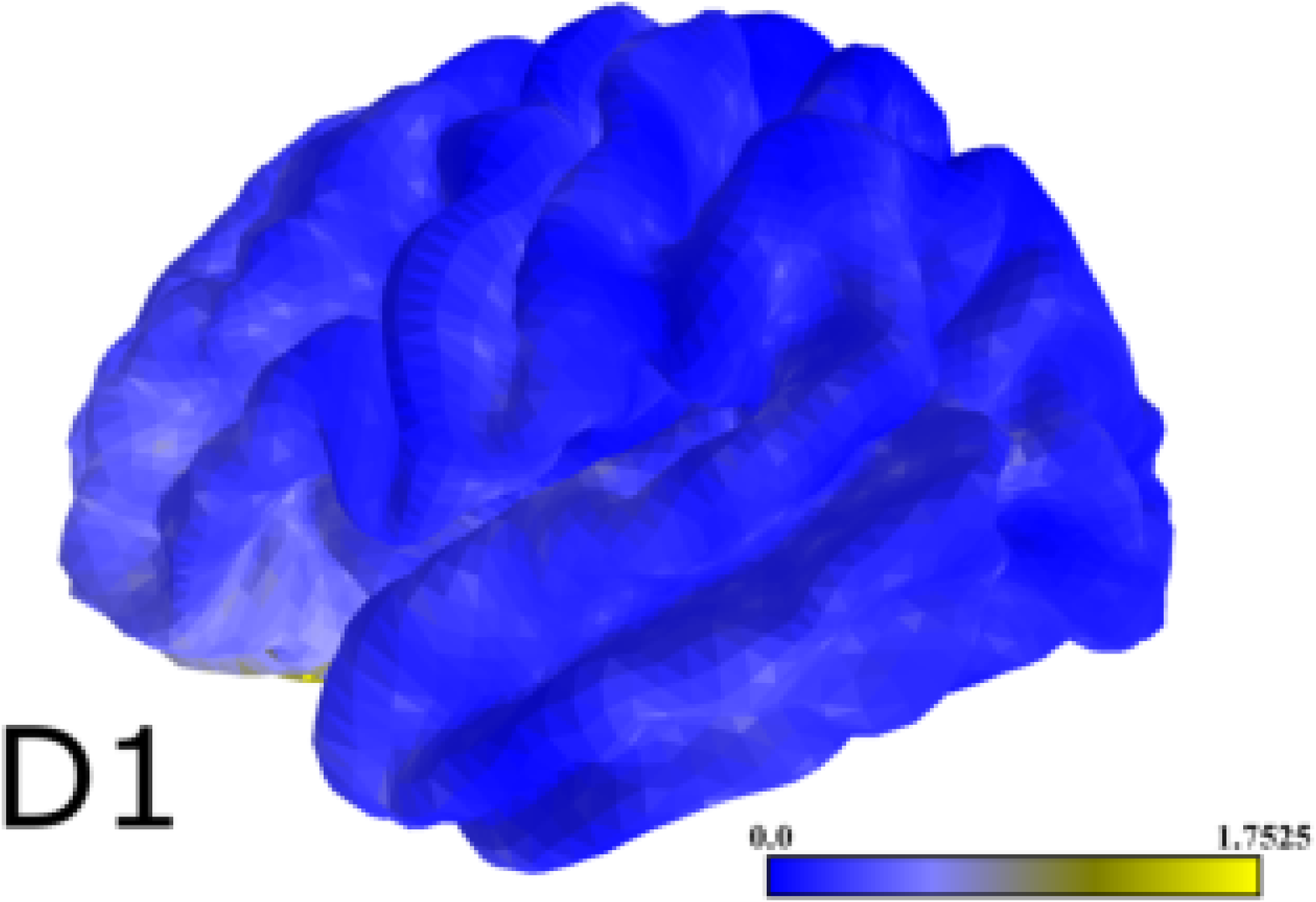}
\includegraphics[width=0.22\linewidth]{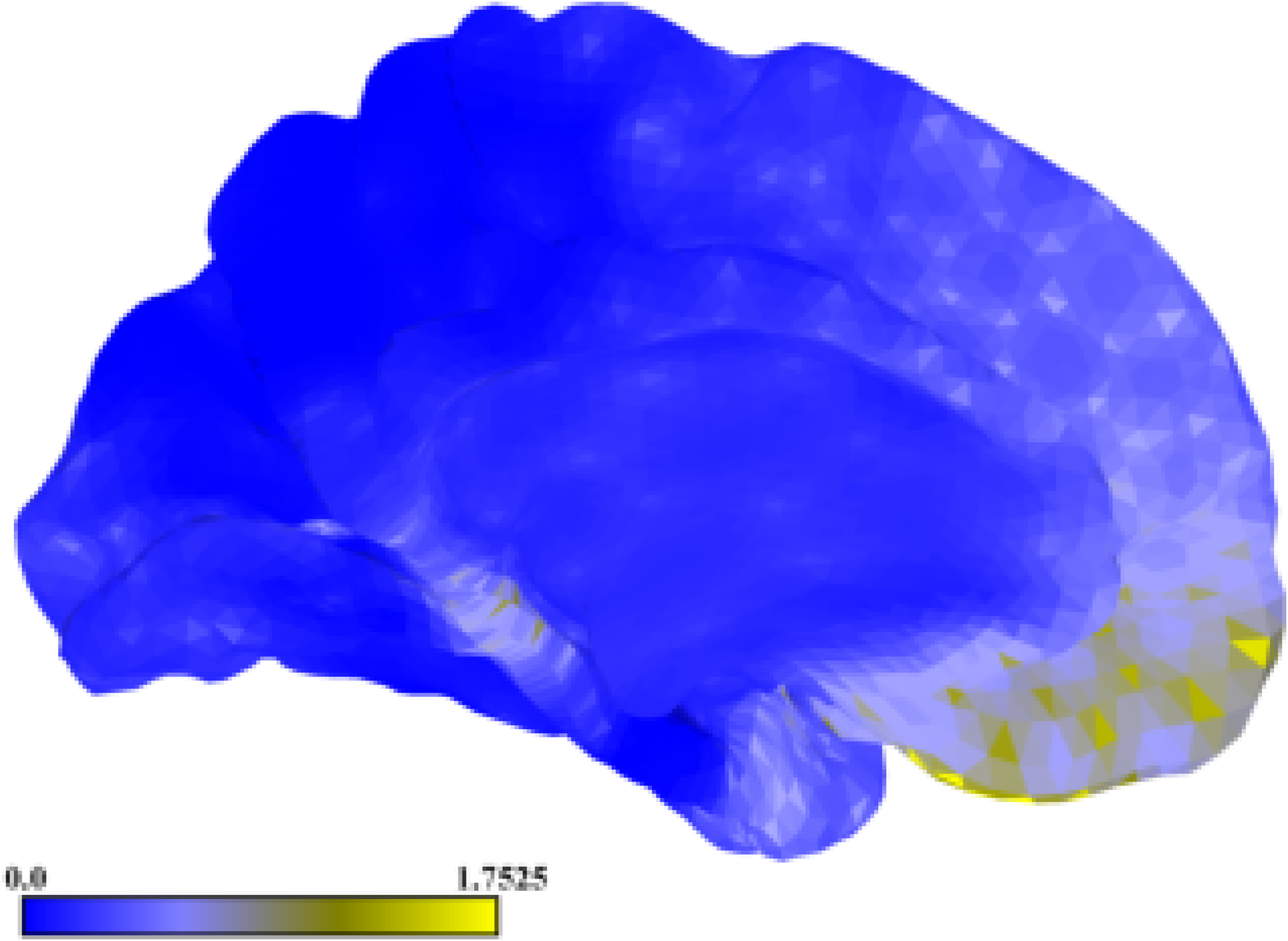} \\
\includegraphics[width=0.22\linewidth]{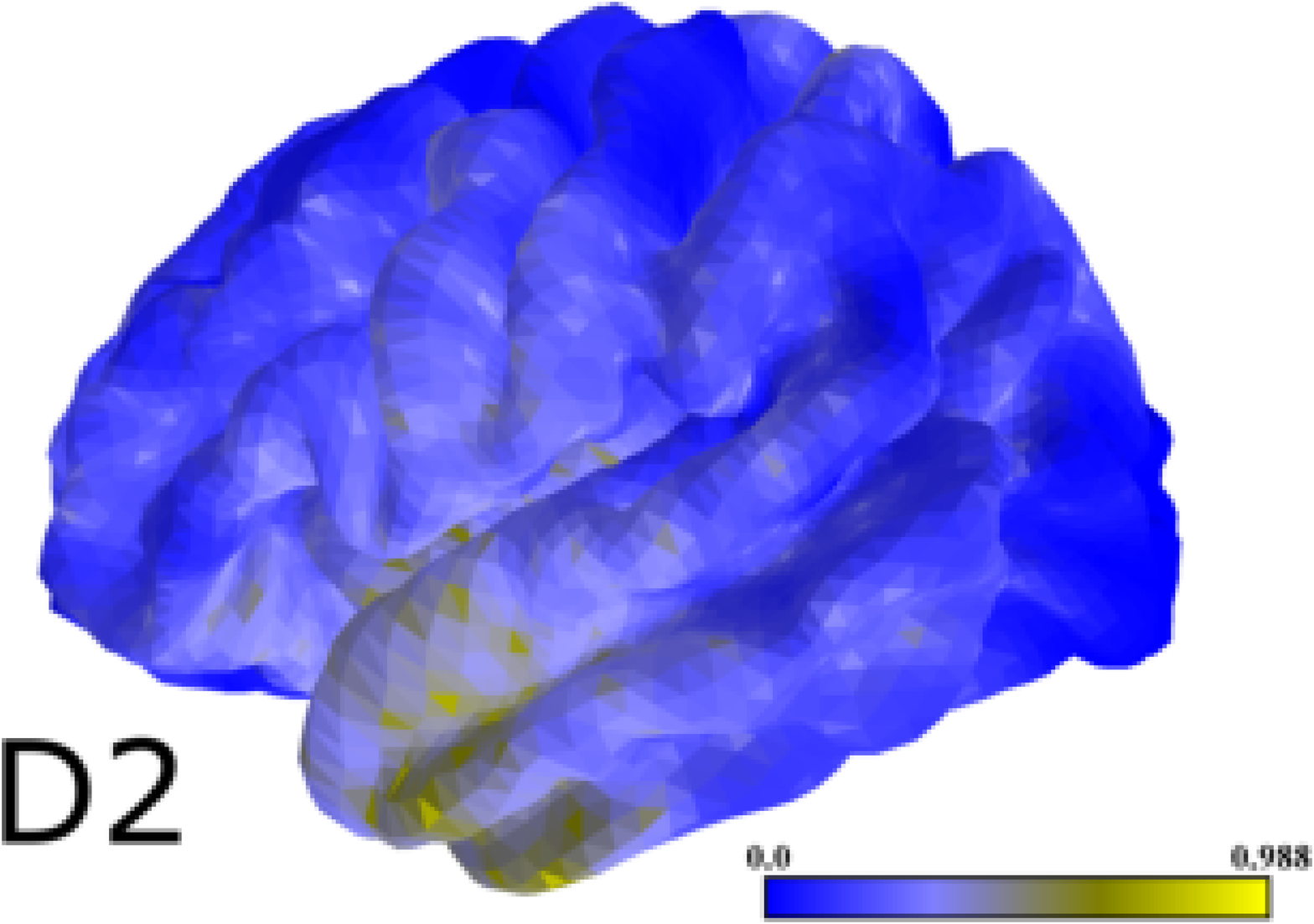}
\includegraphics[width=0.22\linewidth]{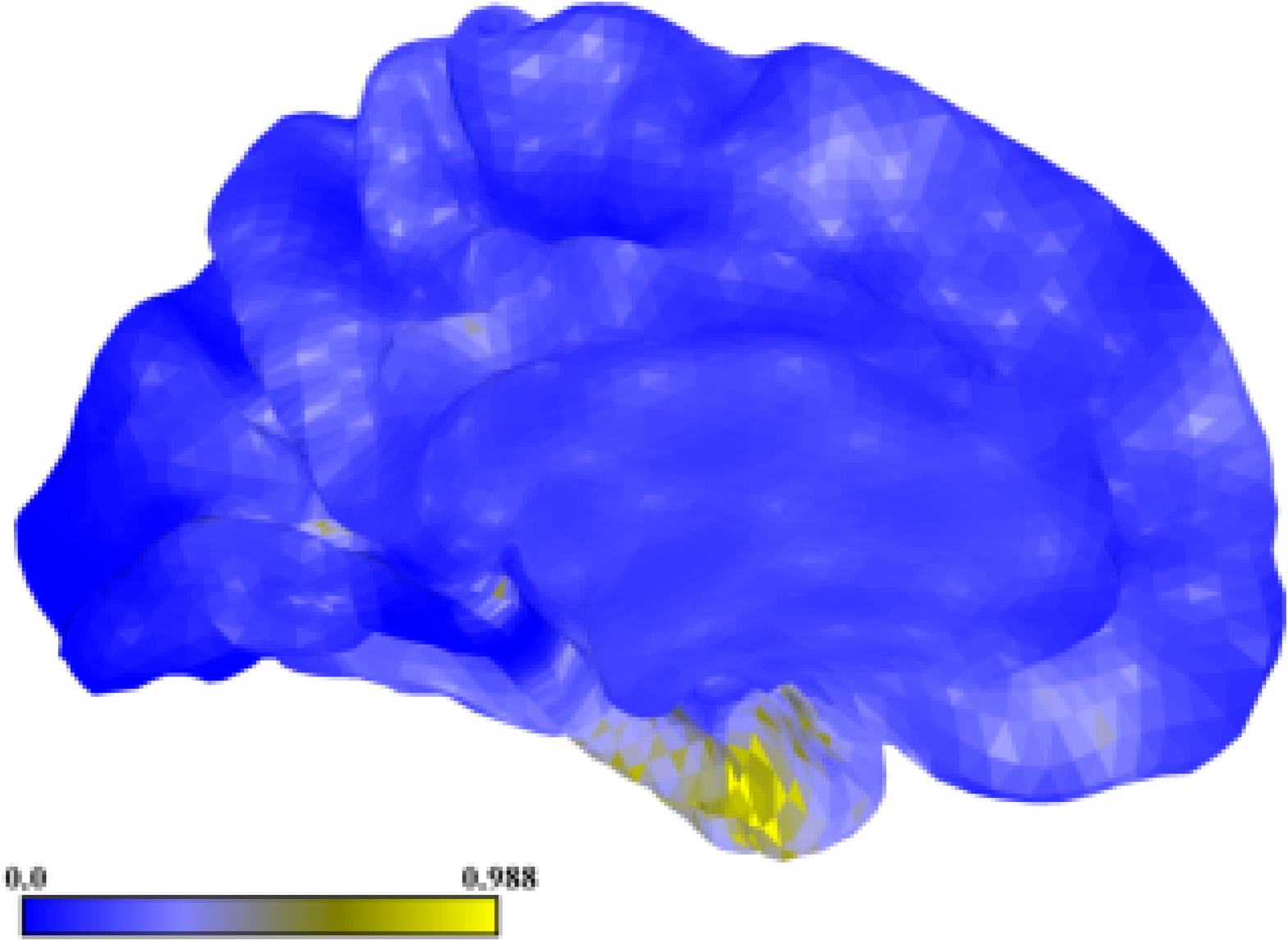}
\includegraphics[width=0.22\linewidth]{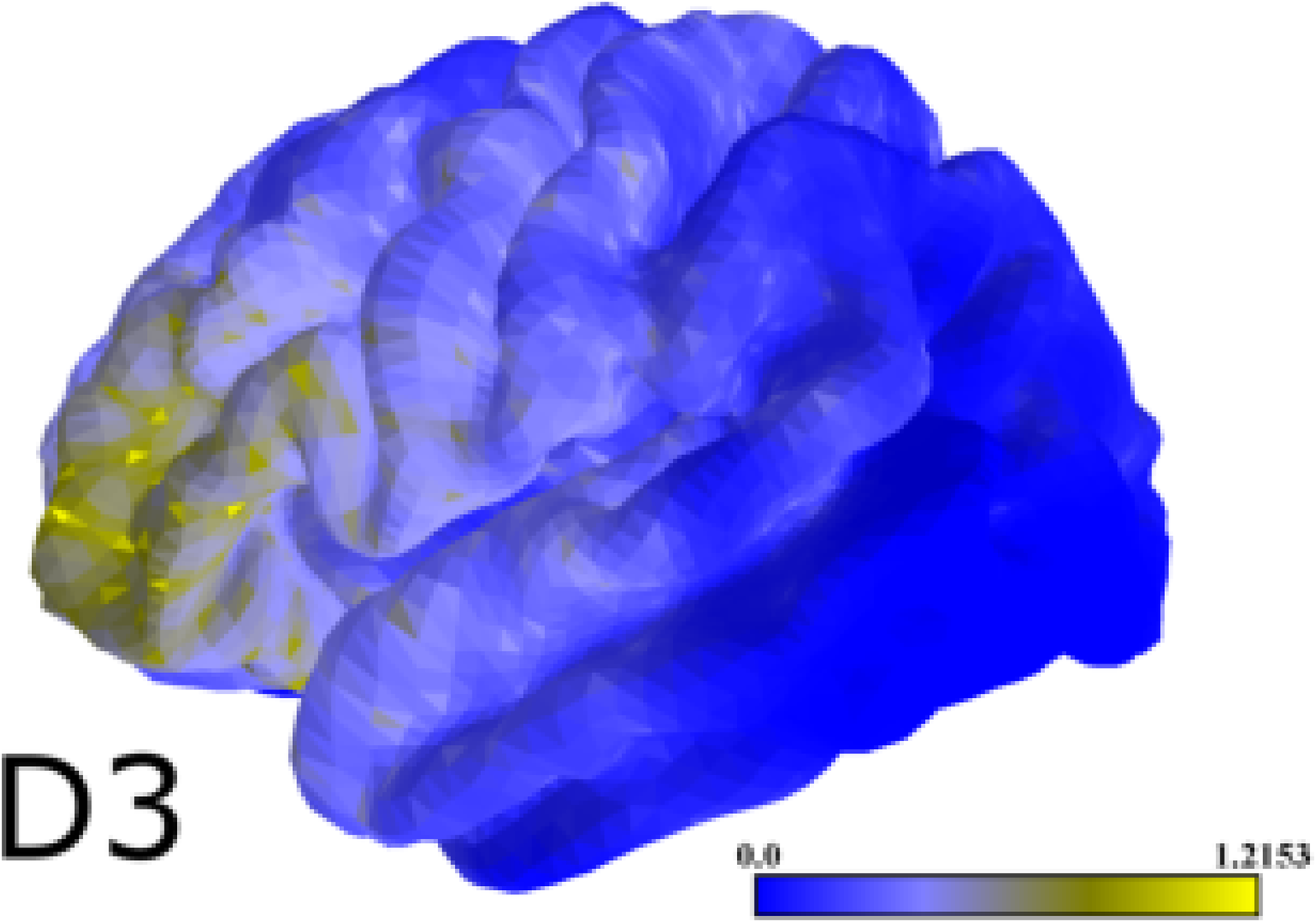} 
\includegraphics[width=0.22\linewidth]{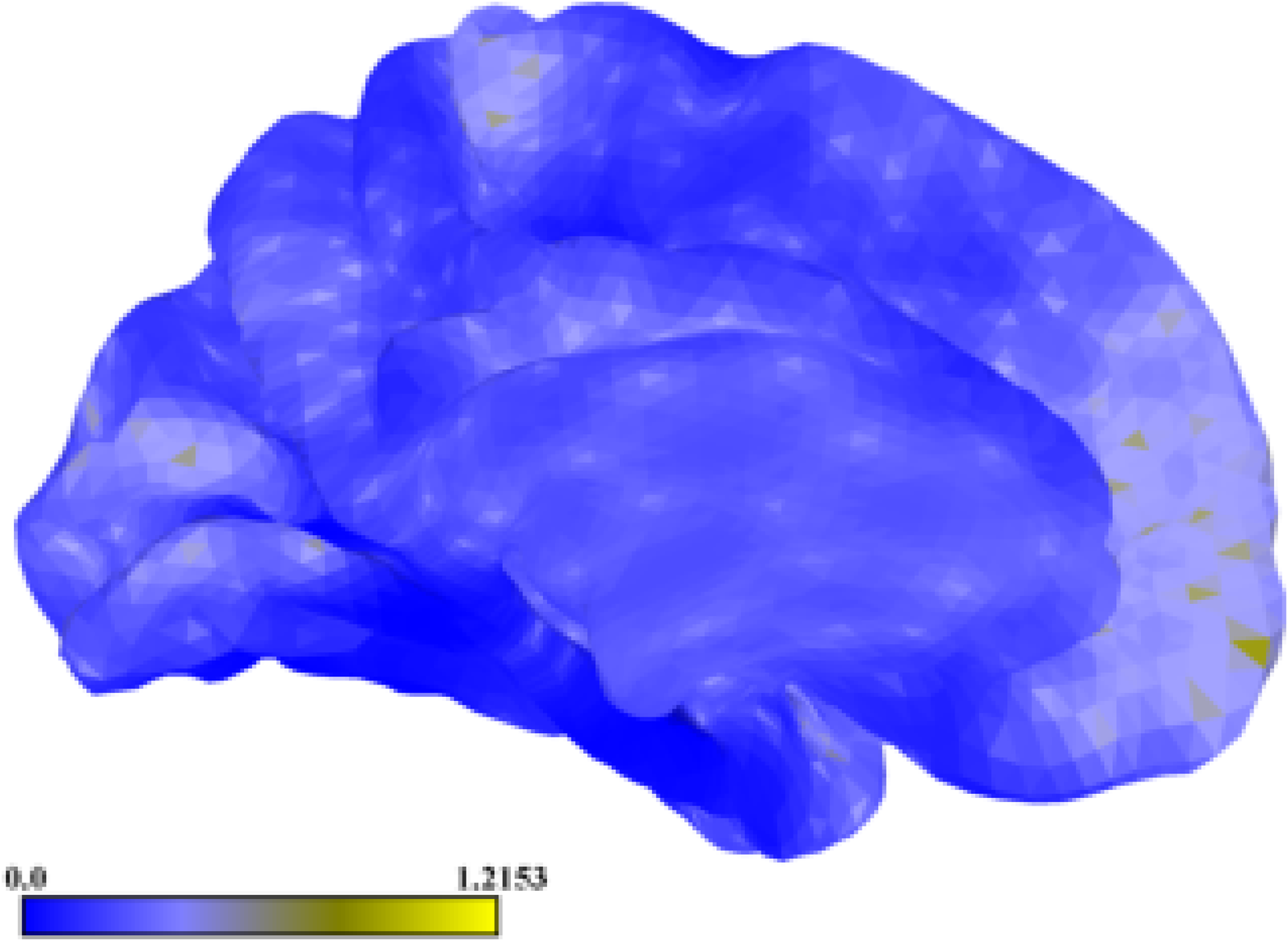}
\caption{A. median myelin. B. myelin standard deviation. When LE-PNNMF is used for factorizing these data: C. number of time a face has been divided during the first 600 refinements and D.1-3 first three basis generated during this optimization.}
\label{res:myelin}
\end{figure}

Figure \ref{res:final} presents the L1 norm of B and C obtained for all our algorithm and datasets. 
Ten steps of refinement were conducted. At each step, five faces were subdivided and the algorithms run for ten optimization steps. 
Our results suggest that the tradeoff between the sparsity of B and C is different for the parameters selected. 
Because projected NMF do not control the sparsity of the loadings, PNMF basis tend to be very sparse but the projected loadings are not. 
The other factorization schemes balanced the L1 norms of B and C. For the parameters selected, the basis generated by dictionary learning were slightly sparser. 
We illustrate in figure \ref{res:myelin} the results obtained when decomposing the myelin data using LE-PNNMF for a larger number of iterations. The basis obtained nicely decompose the map of large data variability into weakly overlapping components. The refinement had focused accordingly.

\section{Discussions}

In this paper we exploit the structure of the icosahedral meshes commonly 
used in neuroimaging for accelerating optimization tasks such as data factorization. 
We compare four factorizations schemes and investigate the use of extrapolation for 
further reducing computational time. 
Our experiments with structural and functional data acquired by the Human Connectome Project 
demonstrate that our approach is particularly interesting for processing massive datasets.

\newpage
\bibliographystyle{splncs03}
\bibliography{biblio,biblio2}

\begin{thebibliography}{10}
\providecommand{\url}[1]{\texttt{#1}}
\providecommand{\urlprefix}{URL }

\bibitem{fista}
{Beck}, A., {Teboulle}, M.: A fast iterative shrinkage-thresholding algorithm
  for linear inverse problems. Siam Journal of Imaging Sciences  2(1),  183 --
  202 (2009)

\bibitem{palm}
{Bolte}, J., {Sabach}, S., {Teboulle}, M.: Proximal alternating linearized
  minimization for nonconvex and nonsmooth problems. Math. Progr.  146(1),  459
  -- 494 (2014)

\bibitem{cichocki_book}
{Cichocki}, A., {Zdunek}, R., {Phan}, A., {Amari}, S.: Nonnegative Matrix and
  Tensor Factorizations: Applications to Exploratory Multi-way Data Analysis
  and Blind Source Separation. Wiley Publishing (2009)

\bibitem{freesurfer}
{Dale}, A., {Fischl}, B., {Sereno}, M.: Cortical surface-based analysis. i.
  segmentation and surface reconstruction. Neuroimage  9,  179 -- 194 (1999)

\bibitem{hcp_preprocessing}
{Glasser}, {Sotiropoulos}, {Wilson}, {Coalson}, {Fischl}, {Andersson}, {Xu},
  {Jbabdi}, {Webster}, {Polimeni}, {Van Essen}, {Jenkinson}: The minimal
  preprocessing pipelines for the human connectome project. Neuroimage  80,
  105 -- 124 (2013)

\bibitem{wavelet_chung}
{Kim}, W., {Singh}, V., {Chung}, M., {Hinrichs}, C., {Pachauri}.D., {Okonkwo},
  O., {Johnson}, S.: Multi-resolutional shape features via non-euclidean
  wavelets: Applications to statistical analysis of cortical thickness.
  NeuroImage  93,  107 -- 123 (2014)

\bibitem{nmf_init}
{Langville}, A., {Meyer}, C., {Albright}.R.: Initializations for the
  nonnegative matrix factorization. In: KDD 2006 (2006)

\bibitem{nnmf}
{Lee}, D., {Seung}, H.S.: Algorithms for nonnegative matrix factorization. In:
  Advances in Neural Information Processing Systems (NIPS). pp. 556 -- 562
  (2000)

\bibitem{wavelet_original}
{Schr\"{o}der}, P., {Sweldens}, W.: Spherical wavelets: Efficiently
  representing functions on the sphere. In: SIGGRAPH'95. pp. 161 -- 172 (1995)

\bibitem{enflure2}
{Sotiras}, A., {Resnick}, S., {Davatzikos}, C.: Finding imaging patterns of
  structural covariance via non-negative matrix factorization. NeuroImage.
  108,  1 -- 16 (2015)

\bibitem{hcp}
{Van Essen}, D., {Smith}, S., {Barch}, D., {Behrens}, T., {Yacoub}, E.,
  {Ugurbil,K. for the {WU-M}inn HCP Consortium}: The {WU-M}inn human connectome
  project: An overview. NeuroImage  80,  62 -- 79 (2013)

\bibitem{caret}
{Van Essen}, D., {Dickson}, J., {Harwell}, J., {Hanlon}, D., {Anderson}, C.,
  {Drury}, H.: An integrated software system for surface-based analyses of
  cerebral cortex. Journal of American Medical Informatics Association  8(5),
  443 -- 459 (2001)

\bibitem{pnmf2}
{Yang}, Z., {Oja}, E.: Linear and nonlinear projective nonnegative matrix
  factorization. IEEE Transactions on Neural Networks  21(5),  734 -- 749
  (2010)

\bibitem{wavelet_golland}
{Yu}.P., {Grant}, P., {Qi}, Y., {Han}, X., {S\'{e}gonne}, F., {Pienaar}, R.,
  {Busa}, E., {Pacheco}, J., {Makris}, N., {Buckner}, R., {Golland}, P.,
  {Fischl}, B.: Cortical surface shape analysis based on spherical wavelets.
  IEEE Tr. on Medical Imaging  26(4),  582 -- 597 (2007)

\bibitem{reho}
{Zang}, Y., {Jiang}, T., {Lu}, Y., {He}, Y., {Tian}, L.: Regional homogeneity
  approach to fmri data analysis. Neuroimage  22,  394 -- 400 (2004)

\bibitem{alff}
{Zang}, Y., {He}, Y., {Zhu}, C., {Cao}, Q., {Sui}, M., {Liang}, M., {Tian}, L.,
  {Jiang}, T., {Wang}, Y.F.: Altered baseline brain activity in children with
  adhd revealed by resting-state functional mri". Brain and Development  29(2),
   83 -- 91 (2007)

\end{thebibliography}

\newpage
\appendix

\section{Proofs}

All the optimization schemes implement an alternative gradient descent. 
They alternatively minimize the basis $B$ and the loadings $C$ by performing a gradient descent, 
which depends on the scheme. 
For the Dictionary Learning, a step of proximal gradient is performed \cite{palm,fista} 
as explained in the next section.

For the non-negative schemes, a multiplicative update is performed \cite{nnmf}. 
This update adapts the step size of the gradient for each matrix component independently, 
so that the matrix components remain positive. 
More precisely, let $Y$ denote $B$ or $C$ and $G$ the gradient of the differentiable part of the 
objective function. $G$ is first decomposed into its positive and negative parts:
\begin{eqnarray}
G=G^{+}-G^{-} ~~~~ G^{+} \geq 0 ~,~ G^{-} \geq 0
\end{eqnarray}
A standard gradient update with positive step size $\alpha$ would be, for any component $Y_{i,j}$:
\begin{eqnarray}
Y_{i,j} \leftarrow Y_{i,j} - \alpha \left(G^{+}_{i,j}-G^{-}_{i,j}\right)
\end{eqnarray}

For nonnegative factorization, the positive step size $\alpha$ is set independently for each component as follows \cite{nnmf}:
\begin{eqnarray}
\alpha_{i,j} = \frac{Y_{i,j}}{G^{+}_{i,j}}
\end{eqnarray}
which yields
\begin{eqnarray*}
Y_{i,j} \leftarrow \frac{Y_{i,j}}{G^{+}_{i,j}}\left(G^{+}_{i,j}-G^{+}_{i,j}+G^{-}_{i,j}\right) = \frac{Y_{i,j}}{G^{+}_{i,j}}G^{-}_{i,j}
\end{eqnarray*}

This update maintains the positivity of $Y$ and can be expressed in a more elegant form using componentwise products and divisions:
\begin{eqnarray}
Y \leftarrow Y \odot G^{-} \oslash  G^{+}
\end{eqnarray}

All the gradients were derived using G\^{a}teaux derivatives, as explained in the next sections. 
We introduce the notation $\langle .,. \rangle$ for the standard inner product, hence:
\begin{eqnarray}
||Y||_2^2=\langle Y,Y \rangle
\end{eqnarray}

We recall the following notations introduced in the paper:
\begin{eqnarray}
L&=&X^{T} D \\ 
K&=&D^{T} D \\ 
M&=&L^{T} L = D^{T} X X^{T} D
\end{eqnarray}

\section{Dictionary Learning}

The differenciable part of the objective of the parametric dictionary learning:
\begin{eqnarray}
&(DL)& ~~~~\min{ \frac{1}{2}||X-DBC||_2^2 + \lambda\left(||B||_1+||C||_1\right) }
\end{eqnarray}
can be expressed as follows:
\begin{eqnarray*}
e=\frac{1}{2}\left[ \langle X,X \rangle+\langle DBC,DBC \rangle-2\langle X,DBC \rangle \right]
\end{eqnarray*}

As a result, for $B$, the G\^{a}teaux derivative of $e$ in the direction $F$ writes:
\begin{eqnarray*}
\frac{\partial e(B+\tau F)}{\partial \tau}|_{\tau=0} &=& \frac{1}{2}\frac{\partial}{\partial \tau}\left[ \langle X,X \rangle+\langle DBC,DBC \rangle-2\langle X,DBC \rangle 
+2\tau\langle DFC,DBC \rangle \right. \\
&+& \left.\tau^2 \langle DFC,DFC \rangle -2\tau\langle X,DFC \rangle \right]|_{\tau=0} \\
 &=& \langle F,D^{T}DBCC^{T} \rangle-\langle F,D^{T}XC^{T} \rangle \\
 &=& \langle F,KDBCC^{T} - L^{T}C^{T} \rangle
\end{eqnarray*}
Hence, a gradient descent of step $\eta$ on $B$ writes:
\begin{eqnarray*}
B \leftarrow B - \eta \left(KDBCC^{T} - L^{T}C^{T}\right)
\end{eqnarray*}
Applying a componentwise soft-thresholding $S$ by $\eta\lambda$ \cite{palm,fista} leads to the optimization step describe in Section 2.1.

The derivation for $C$ is very similar:
\begin{eqnarray*}
\frac{\partial e(C+\tau F)}{\partial \tau}|_{\tau=0} &=& \frac{1}{2}\frac{\partial}{\partial \tau}\left[ \langle X,X \rangle+\langle DBC,DBC \rangle-2\langle X,DBC \rangle 
+2\tau\langle DBF,DBC \rangle \right. \\
&+& \left.\tau^2 \langle DBF,DBF \rangle -2\tau\langle X,DBF \rangle \right]|_{\tau=0} \\
 &=& \langle F,B^{T}D^{T}DBC \rangle-\langle F,B^{T}D^{T}X \rangle \\
 &=& \langle F, B^{T}KBC-B^{T}L^{T} \rangle
\end{eqnarray*}

As a result, our parametric dictionary learning writes:
\begin{eqnarray*}
&(DL)& ~~~~\min{ \frac{1}{2}||X-DBC||_2^2 + \lambda\left(||B||_1+||C||_1\right) }\\
&~&\texttt{by repeating~~}
\left\{\begin{array}{l}
B \leftarrow S\left(B-\eta \left(KBCC^{T}-L^{T}C^{T}\right),\lambda \eta \right) \\
C \leftarrow S\left(C-\eta \left(B^{T}KBC-B^{T}L^{T}\right),\lambda \eta \right) 
\end{array}\right.
\end{eqnarray*}

\section{PPNMF}

For our parametric PNMF scheme \cite{pnmf2}
\begin{eqnarray*}
(PPNMF) ~~~~\min{ ||X-DBB^{T}D^{T}X||_2^2 } ~~~~~~~~ B \geq 0 ~~~~~~~~~~~~~~~~~~~~~~~~\\
\end{eqnarray*}
the G\^{a}teaux derivative in the direction $F$ writes:
\begin{eqnarray*}
\frac{\partial e(B+\tau F)}{\partial \tau}|_{\tau=0} &=& 
-2\langle X,DFB^{T}D^{T}X \rangle+2\langle DFB^{T}D^{T}X,DBB^{T}D^{T}X \rangle \\
&~&-2\langle X,DBF^{T}D^{T}X \rangle+2\langle DBF^{T}D^{T}X,DBB^{T}D^{T}X \rangle \\
&=&-2\langle F,D^{T}XX^{T}DB \rangle+2\langle F,D^{T}DBB^{T}D^{T}XX^{T}DB \rangle \\
&~&-2\langle XX^{T}DF,DB \rangle+2\langle DB,DBB^{T}D^{T}XX^{T}DF \rangle \\
&=& \langle F,-4D^{T}XX^{T}DB \rangle+\langle F,2D^{T}DBB^{T}D^{T}XX^{T}DB \rangle \\
&~&+\langle F,2D^{T}XX^{T}DBB^{T}D^{T}DB \rangle \\
&=& \langle F,-4MB+2\left(KBB^{T}M+MBB^{T}K\right)B \rangle \\
\end{eqnarray*}

which leads to the following multiplicative update:
\begin{eqnarray*}
B \leftarrow B \odot 2MB \oslash \left[ \left(KBB^{T}M+MBB^{T}K\right)B \right]
\end{eqnarray*}

Shrinking the amplitude of this update by two as follows:
\begin{eqnarray*}
B \leftarrow \frac{1}{2}B+ \frac{1}{2}\left[B \odot 2MB \oslash \left[ \left(KBB^{T}M+MBB^{T}K\right)B \right]\right]
\end{eqnarray*}
leads to the update presented in Section 2.1: ~\\
\begin{eqnarray*}
(PPNMF) ~~~~\min{ ||X-DBB^{T}D^{T}X||_2^2 } ~~~~~~~~ B \geq 0 ~~~~~~~~~~~~~~~~~~~~~~~~\\
\texttt{by repeating}~~~
B \leftarrow B \odot \left(\frac{\mathbf{1}(n_k,n_d)}{2}+\left[MB\right] \oslash \left[ \left(KBB^{T}M+MBB^{T}K\right)B \right]\right) \\
\end{eqnarray*}

\section{PNNMF}

For the parametric NMF scheme
\begin{eqnarray*}
&(PNNMF)& ~~~~\min{ ||X-DBC||_2^2 + \lambda\left(||B||^2_2+||C||^2_2\right) }~~~~ B \geq 0 ~~ C \geq 0\\
\end{eqnarray*}
the G\^{a}teaux derivative with respect to $B$ in the direction $F$ writes:
\begin{eqnarray*}
\frac{\partial e(B+\tau F)}{\partial \tau}|_{\tau=0} &=& 
2\langle DBC,DFC \rangle - 2\langle X,DFC \rangle +2\lambda \langle F,B \rangle \\
 &=& 
2\langle F,D^{T}DBCC^{T} \rangle - 2\langle F,D^{T}XC^{T} \rangle +2\lambda \langle F,B \rangle \\
&=& 
2\langle F,KBCC^{T} - L^{T}C^{T} +\lambda B \rangle \\
\end{eqnarray*}
Following \cite{nnmf,cichocki_book}, we join the gradient term originating from the penalty with the negative part of the gradient, and we project the components of the matrix back to positive real numbers. These operations yield the following multiplicative update:
\begin{eqnarray*}
B \leftarrow B \odot \left[ L^{T}C^{T}-\lambda B \right]_{+} \oslash \left[KBCC^{T}\right] \\
\end{eqnarray*}

The same derivation yields for $C$:
\begin{eqnarray*}
\frac{\partial e(C+\tau F)}{\partial \tau}|_{\tau=0} &=& 
2\langle DBC,DBF \rangle - 2\langle X,DBF \rangle +2\lambda \langle F,C \rangle \\
 &=& 
2\langle F,B^{T}D^{T}DBC \rangle - 2\langle F,B^{T}D^{T}X \rangle +2\lambda \langle F,C \rangle \\
 &=& 
2\langle F,B^{T}KBC - B^{T}L^{T} +\lambda C \rangle \\
\end{eqnarray*}
hence the update:
\begin{eqnarray*}
C \leftarrow C \odot \left[ B^{T}L^{T}-\lambda C \right]_{+} \oslash \left[B^{T}KBC\right] \\
\end{eqnarray*}

and we obtain the PNNMF scheme presented in Section 2.1.

\section{SPNNMF}

The derivation of the SPNNMF scheme follows exactly the derivation of the PNNMF. 
Since $B$ and $C$, the derivative of the L1 norms is a matrix of ones of the same size
as the updated matrix.

\end{document}